\documentclass{article}
\PassOptionsToPackage{numbers, compress}{natbib}
\usepackage[main,final]{neurips_2026}
\usepackage{amsmath}
\usepackage{multirow}
\usepackage[table]{xcolor}
\usepackage{url}
\usepackage{amsmath,amssymb,amsfonts}
\usepackage{algorithmic}
\usepackage{graphicx}
\usepackage{textcomp}
\usepackage{wrapfig}
\usepackage{caption}
\usepackage{paralist}
\usepackage{longtable}
\usepackage{xspace}
\usepackage{listings}
\usepackage{booktabs}
\usepackage{hhline}
\usepackage{framed,multirow}
\usepackage{lipsum}  
\usepackage{soul}
\usepackage{cancel}
\usepackage{latexsym}
\usepackage{array}
\usepackage{stackengine}
\usepackage{relsize}
\usepackage{multirow}
\usepackage[ruled,vlined]{algorithm2e}
\usepackage[dvipsnames]{xcolor}
\DeclareMathOperator*{\argmax}{arg\,max}

\usepackage[utf8]{inputenc} % allow utf-8 input
\usepackage[T1]{fontenc}    % use 8-bit T1 fonts
\usepackage{hyperref}       % hyperlinks
\usepackage{url}            % simple URL typesetting

\usepackage{amsfonts}       % blackboard math symbols
\usepackage{nicefrac}       % compact symbols for 1/2, etc.
\usepackage{microtype}      % microtypography
\usepackage{xcolor}         % colors
\newcommand{\ours}{\textsc{Conch-Z}\xspace}
\newcommand{\conchlin}{$\textsc{CONCH}_{\text{lin}}$}
\newcommand{\ourslin}{$\ours_{\text{lin}}$}

\newcommand{\nslides}{335,645\xspace}
\title{Beyond Sparse Captions: Aligning Slide-Level Text and Patch-Level Vision in Pathology}

\author{%
  Daniel Shao$^*$ \\
  MIT \\ MGB AI Institute \\ Harvard University \\
  \texttt{dshao@mit.edu} \\
  \And
  Dongmin Bang$^*$ \\
  MGB AI Institute \\ Harvard University \\
  \And
  Luca L. Weishaupt \\
  MIT \\ MGB AI Institute \\ Harvard University \\
  \And
  Sophia J. Wagner \\
  MGB AI Institute \\ Harvard University \\
  \And
  Nic G. Reitsam \\
  University of Augsburg \\ MGB AI Institute \\ Harvard University \\
  \And 
  Ming Y. Lu \\
  MGB AI Institute \\ Harvard University \\
  \And
  Long P. Le \\
  MGB AI Institute \\ Harvard University \\
  \AND
  Richard J. Chen$^\dagger$ \\
  MGB AI Institute \\ Harvard University \\
  \And
  Faisal Mahmood$^\dagger$ \\
  MGB AI Institute \\ Harvard University \\
  \AND
  \textnormal{$^*$Co-first authors} \\
  \textnormal{$^\dagger$Co-senior authors}
}

\begin{document}

\maketitle

\begin{abstract}

Contrastive vision-language (VL) pretraining in computational pathology has generally relied on curated image-caption datasets sourced from social media, educational videos, and research publications. These corpora are sparse relative to the diversity of tissue morphology and difficult to scale further, as pathologists do not routinely produce descriptive captions in clinical practice. We propose to instead leverage slide-level pathology reports which can be generated during routine diagnostic workflows in a scalable manner. A key challenge in this weakly-supervised setting is that pathology reports describe macroscopic, diagnostic findings while patch-level visual features describe microscopic findings. We introduce Locked Image Multiple Instance PreTraining (LIMIT), which addresses this gap by freezing a pretrained vision encoder and fine-tuning only the text encoder: report embeddings serve as cross-attention queries over bags of patch features, producing report-conditioned slide representations optimized via InfoNCE. Using LIMIT, we establish CONCH-Z, which aligns the text encoder of CONCH v1.5 to 335k clinical reports. Evaluated across 22 tasks spanning patch-level classification, tumor detection, and slide-level subtyping, CONCH-Z establishes state-of-the-art zero-shot performance over VL encoders trained on curated caption datasets, while simultaneously closing the gap with slide-level foundation models trained with substantially more parameters and compute. The code and model is available at \url{https://github.com/mahmoodlab/LIMIT}.
\end{abstract}

\section{Introduction}
\label{sec:intro}

Foundation models have advanced computational pathology (CPath), enabling scalable analysis of digitized tissue sections for diagnosis, prognosis, and biomarker prediction \cite{song_artificial_2023, lipkova_deep_2022}. In particular, vision-language (VL) models have extended patch-level representation learning by aligning visual features with medical text, unlocking multimodal capabilities such as zero-shot classification, text-to-image retrieval, and slide-level inference without task-specific supervision \cite{lu_visual-language_2024, xiang_visionlanguage_2025, lu_visual_2023}. VL encoders have consequently become integral to a growing number of CPath systems, serving as upstream modules for slide-level architectures \cite{ding_multimodal_2025, huang_act_2026}, explainability frameworks \cite{shao_mixture_2025_v2}, and reasoning systems \cite{chen_slidechat_2025, liang_wsi-llava_2025}.

In current foundation models development, while vision-only pretraining recipes have naturally scaled to pathology via abundance of unlabeled tissue patches \cite{chen_towards_2024, vorontsov_virchow_2024}, VL pretraining faces a fundamental data bottleneck. General-purpose VL models such as CLIP and CoCa \cite{radford_learning_2021, yu2022coca} are pretrained on hundreds of millions of image-text pairs curated from the open web. By contrast, pathology VL models must rely on far smaller, domain-specific corpora that are difficult to scale, sourced from social media platforms, educational videos, and open-access research publications \cite{huang_visuallanguage_2023, ikezogwo_quilt-1m_nodate, lu_visual-language_2024}. While these sources have enabled the initial development of pathology VL encoders, they do not reflect standard clinical practice, as pathologists do not routinely produce descriptive morphological captions outside of academic or educational contexts. The resulting datasets, typically on the order of one million curated pairs, are sparse relative to the diversity of tissue morphology, linguistically narrow, and difficult to scale further within the same sourcing paradigm. Continuing to improve VL alignment in pathology therefore requires moving beyond curated captions towards the scalable generation of slide-level pathology reports from routine diagnostic workflows as a supervisory signal.

%\begin{figure}
%    \centering
%    \includegraphics[width=1\linewidth]{figures/fig1.png}
%    \caption{\textbf{Slide-level classification for cancer subtyping}. Overall performance on 1,348 slides from TCGA for multi-organ cancer subtyping. Zero-shot aggregation with linear probe achieves comparable performance to supervised aggregators trained on 8,226 samples. Pairing zero-shot aggregation and classification with \ours yields the highest performance by a wide margin.}
%    \label{fig:fig1}
%\end{figure}
This work addresses the challenge of vision-language alignment under weak supervision, where slide-level reports describe macroscopic diagnostic findings but only patch-level visual features are available. This setting mirrors the classical MIL formulation in CPath, which learns to aggregate patch representations for slide-level prediction without patch-level labels, but extended to the cross-modal case, where the supervisory signal is an unstructured text report rather than a class label.

To this end, we propose \textbf{L}ocked \textbf{I}mage \textbf{M}ultiple \textbf{I}nstance Pre\textbf{T}raining (LIMIT), which freezes a pretrained vision encoder and fine-tunes only the text encoder by using report embeddings as cross-attention queries over bags of patch embeddings, producing report-conditioned slide representations that can be optimized via InfoNCE in slide classification pretext tasks. Using LIMIT, we present \ours, which aligns the text encoder of CONCHv1.5\cite{lu_visual-language_2024} to \nslides clinical pathology reports for improved zero-shot cross-modal capabilities. Evaluated across 22 diverse benchmarks, \ours reaches SOTA performance on zero-shot patch-level classification compared to VL encoders trained on curated caption datasets, while simultaneously closing the gap with supervised MIL slide-level foundation models (\textit{e.g.}, PRISM, TITAN) trained with orders of magnitude more parameters and compute (\textbf{Table \ref{tab:slide_zeroshot}}). Our main contributions are as follows:

\begin{itemize}
\item We formalize vision-language pretraining for pathology as a weakly-supervised cross-modal MIL problem and introduce LIMIT, a method that fine-tunes only the text encoder via report-guided cross-attention over frozen patch embeddings.
\item We train \ours on \nslides clinical reports and establish state-of-the-art zero-shot performance across diverse benchmarks, including pan-cancer subtyping, tumor detection, and patch-level classification.
\item We show that slide-level report supervision, mediated through text-guided attention, strengthens fine-grained visual-semantic alignment, yielding simultaneous improvements at both the patch and slide level without requiring local annotations or learned aggregation parameters.

\end{itemize}

\section{Related Works}
\label{sec:related_works}

\textbf{Vision-Language Models in Computational Pathology.}
Vision-language (VL) modeling has shifted computational pathology from task-specific prediction toward general-purpose multimodal foundation models. Early patch-level VL models such as PLIP aligned pathology images with text from medical social media and academic sources \cite{huang_visuallanguage_2023}, while CONCH and Quilt-1M scaled this paradigm to over one million curated image-caption pairs from PubMed Central, educational sources, YouTube, and social media \cite{lu_visual-language_2024, ikezogwo_quilt-1m_nodate}. MUSK further expanded pretraining with 50 million pathology images and 1 billion unpaired text tokens, followed by contrastive alignment on Quilt-1M \cite{xiang_visionlanguage_2025}. However, these patch-level VL encoders remain primarily trained on curated captions rather than the slide-level diagnostic reports routinely produced in clinical practice.

Several recent works instead leverage slide-level supervision from whole-slide images and pathology reports. PRISM, Prov-GigaPath, PathAlign, and TITAN learn slide-level representations by training dedicated aggregation modules or slide encoders on top of patch features, using report supervision, synthetic captions, or large-scale self-supervised objectives \cite{shaikovski_prism_nodate, xu_whole-slide_2024, ahmed_pathalign_2024, ding_multimodal_2025}. These models establish the value of slide-report pairs for whole-slide representation learning, but their compressed slide-level embeddings can reduce native access to fine-grained patch-level visual-language alignment, which is important for patch preprocessing \cite{wu_learning_2025, huang_act_2026} and post-hoc interpretability \cite{shao_mixture_2025_v2}. In contrast, LIMIT uses slide-level clinical reports to adapt a patch-level VL model, aiming to retain unified patch- and slide-level zero-shot capabilities without training a new slide encoder.

\textbf{Cross-Attention Pooling in Vision-Language Models.}
Cross-attention is widely used to aggregate visual features conditioned on text or learned queries. CoCa uses cross-attention over ViT patch tokens \cite{yu2022coca}, BLIP-2's Q-Former queries frozen visual features \cite{li2023blip}, and X-Pool uses text embeddings as queries over video frame features for text-video retrieval \cite{gorti2022x}. In computational pathology, MIL methods such as ABMIL, CLAM, and TransMIL learn attention-based aggregation over patch embeddings for slide-level prediction \cite{ilse_attention-based_2018, lu_data-efficient_2021, shao_transmil_2021}. LIMIT differs by deriving the attention query from the pathology report text encoder and using this report-conditioned pooling directly as the alignment objective for locked-image text tuning.

\textbf{Locked-image Tuning and Modality Alignment.}
Locked-image Tuning (LiT) freezes a pretrained vision encoder and updates only the text encoder, aligning language representations to an already robust visual space \cite{zhai_lit_2022}. This is especially useful in pathology, where fine-tuning the visual backbone on noisy report supervision may risk degrading learned morphological features. LIMIT applies this principle to \nslides slide-level clinical reports, aligning the text encoder of a patch-level VL model to frozen pathology image features through weak slide-level supervision.

\label{sec:methods}
\begin{figure}[t!]%htbp!]
    \centering
    \includegraphics[width=1\linewidth]{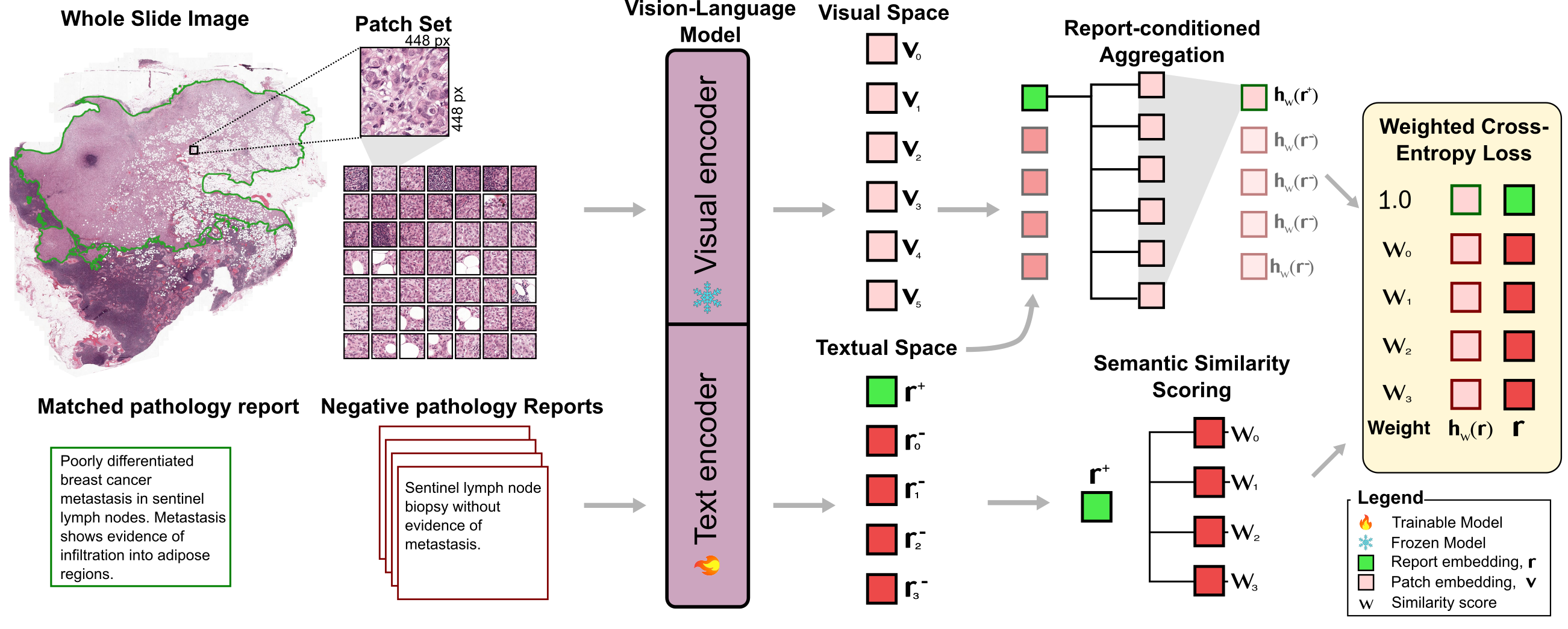}
    \caption{\textbf{Overview of LIMIT.} A frozen vision encoder extracts patch embeddings from the WSI. The text encoder embeds each candidate report as a cross-attention query over the patch bag, producing a report-conditioned slide representation. Only the text encoder is updated via a difficulty-weighted InfoNCE objective.}
    \label{fig:architecture}
\end{figure}

\section{Methods}

% In this section, we outline our approach for vision-language alignment in whole slide images (WSIs). We employ a locked-image text-tuning strategy, wherein the image encoder remains frozen to preserve foundational visual features while the text encoder is actively fine-tuned. To handle the massive resolution of WSIs, we introduce a text-guided attention mechanism to dynamically aggregate patch-level features into a report-conditioned slide-level representation. Finally, we optimize the text encoder using a difficulty-weighted multi-class cross-entropy objective that leverages inter-report semantic similarities to scale the influence of hard negatives.

In this section, we describe Locked Image Multiple Instance PreTraining (LIMIT), our approach for weakly-supervised vision-language alignment in whole slide images (WSIs). LIMIT employs a locked-image tuning strategy, wherein the image encoder remains frozen while the text encoder is fine-tuned. To bridge the spatial disconnect between slide-level reports and patch-level visual features, LIMIT uses report embeddings as cross-attention queries to dynamically aggregate patch-level features into report-conditioned slide-level representations. The text encoder is optimized using a difficulty-weighted InfoNCE objective that leverages inter-report semantic similarities to scale the influence of hard negatives.

\subsection{Report-Guided Slide-Level Representation}
\label{subsec:slide_representation}

A whole slide image $W$ is divided into a set of $M$ non-overlapping patches, $P = \{p_1, p_2, \dots, p_M\}$. We extract patch embeddings using a frozen, pretrained vision encoder $E_I$, yielding a set of visual embeddings $\mathcal{V} = \{\mathbf{v}_1, \dots, \mathbf{v}_M\}$, where $\mathbf{v}_i \in \mathbb{R}^d$.

To aggregate these dense patch representations into a single slide-level embedding, LIMIT introduces a text-driven attention mechanism conditioned on the pathology reports themselves. For a given pathology report $R$, we embed it using our trainable text encoder $E_T$ to obtain the report vector $\mathbf{r} = E_T(R) \in \mathbb{R}^d$. We then compute the alignment between this report embedding and each patch embedding via a scaled dot-product similarity.

Because the aggregation is guided by the text, the resulting slide-level representation $\mathbf{h}_W(\mathbf{r}) \in \mathbb{R}^d$ becomes uniquely conditioned on the specific report being evaluated. It is computed as the softmax-weighted sum of the patch embeddings within the bag:
\begin{equation}
    \alpha_i(\mathbf{r}) = \frac{\exp(\mathbf{v}_i^\top \mathbf{r} / \tau_{attn})}{\sum_{k=1}^M \exp(\mathbf{v}_k^\top \mathbf{r} / \tau_{attn})}
\end{equation}
\begin{equation}
    \mathbf{h}_W(\mathbf{r}) = \sum_{i=1}^M \alpha_i(\mathbf{r}) \mathbf{v}_i
\end{equation}
where $\tau_{attn}$ is a temperature hyperparameter. This dynamic aggregation ensures that the final slide representation natively attends to the morphological regions within the WSI bag that are most relevant to the provided clinical text.

\subsection{Difficulty-Weighted Contrastive Optimization}
  \label{subsec:loss}
 
  During training, each WSI $W$ is paired with its corresponding ground-truth pathology report $R^+$. To construct a robust learning objective, we sample $N$ additional slide-report pairs exclusively from the
  same organ type to serve as hard negatives. Together with the anchor pair, these form a bag of $K=N+1$ paired examples, $\mathcal{B}=\{(W_i,R_i)\}_{i=1}^{K}$. Each report is embedded using the text encoder to
  obtain $\mathbf{r}_i=E_T(R_i)$. Every slide in the bag is matched against all $K$ reports, with its own report serving as the positive and the remaining reports serving as negatives.
 
  Because negative reports sampled from the same organ may exhibit high degrees of semantic overlap with the corresponding positive report (e.g., identical cancer subtype and grade), treating all negatives
  uniformly risks penalizing the model for pushing apart semantically valid alternative descriptions. To address this, we compute the cosine similarity between each positive report and its negative reports to
  dynamically downweight the penalty applied to semantically overlapping negatives. Letting $\text{sim}(\cdot, \cdot)$ denote this similarity, we define a linear-scale weighting factor $w_{ij}$ for each
  mismatched pair:
  \begin{equation}
  w_{ij} = 1 - \frac{\text{sim}(\mathbf{r}_i, \mathbf{r}_j) + 1}{2},
  \qquad i \neq j.
  \end{equation}
  Importantly, gradients do not flow through this weighting component.
 
  The model is optimized as a multi-class classification task, predicting the correct report out of the $K$ candidates for every slide in the bag. Because our slide embeddings are report-dependent, we generate a
  unique, text-conditioned slide representation $\mathbf{h}_{W_i}(\mathbf{r}_j)$ for each slide-report combination. We compute the similarity between these conditioned representations and their respective text
  embeddings. The final training objective is a similarity-weighted InfoNCE loss averaged over all slides in the bag:
  \begin{equation}
      \mathcal{L} = -\frac{1}{K}\sum_{i=1}^{K}
      \log \frac{\exp(\mathbf{h}_{W_i}(\mathbf{r}_i)^\top \mathbf{r}_i / \tau_{nce})}
      {\exp(\mathbf{h}_{W_i}(\mathbf{r}_i)^\top \mathbf{r}_i / \tau_{nce})
      + \sum_{\substack{j=1\\j\neq i}}^{K}
      w_{ij}\exp(\mathbf{h}_{W_i}(\mathbf{r}_j)^\top \mathbf{r}_j / \tau_{nce})}.
  \end{equation}
  where $\tau_{nce}=0.07$ is the contrastive temperature scalar. By freezing the image encoder and backpropagating $\mathcal{L}$ exclusively through $E_T$, LIMIT forces the text encoder's representation space to
  align with the locked visual space, ensuring the text embeddings can effectively localize the nuanced clinical heterogeneity described in the reports.

\subsection{Inference and Evaluation}
\label{subsec:evaluation}

During inference and evaluation, the resulting model finetuned by LIMIT is tasked with selecting the most appropriate text from a set of $K$ candidate options (e.g., reference reports or diagnostic categories), denoted as $\mathcal{C} = \{c_1, c_2, \dots, c_K\}$. 

First, we encode each candidate option with the fine-tuned text encoder to obtain a set of text embeddings, $\mathbf{c}_k = E_T(c_k)$ for $k \in \{1, \dots, K\}$. Given a test whole slide image $W$, we then extract its visual patch embeddings using the frozen vision encoder $E_I$ and aggregate them into a slide-level feature vector $\mathbf{h}_W$ using the same query-guided attention mechanism described in Section \ref{subsec:slide_representation}. This yields $K$ distinct slide-level representations, each capturing the aggregated slide features conditioned on the semantics of a specific candidate option

To determine the best match, we compute the total similarity between the aggregated slide-level representation $\mathbf{h}_W$ and each candidate text embedding $\mathbf{c}_k$. The model's final prediction $\hat{y}$ is formulated as a retrieval task, selecting the candidate option that maximizes the similarity score:
\begin{equation}
    \hat{y} = \argmax_{k \in \{1, \dots, K\}} \text{sim}(\mathbf{h}_W, \mathbf{c}_k)
\end{equation}
where $\text{sim}(\cdot, \cdot)$ represents the cosine similarity. Because the text encoder has been aligned to the locked visual space during training, this inference formulation allows the framework to evaluate directly on unseen WSI-report pairs and perform zero-shot classification without requiring task-specific linear probing or further downstream fine-tuning.

\subsection{Experiments}
\label{sec:dataset_setup}

\textbf{Datasets.} We evaluate \ours on five datasets spanning 22 slide- and patch-level zero-shot tasks; full dataset details are provided in \textbf{Section~\ref{sec:appendix_dataset_full}}. For slide-level evaluation, we use \textbf{TCGA-OT}~\cite{ding_multimodal_2025}, a cancer subtyping benchmark with 46 OncoTree disease labels across multiple anatomical sites, which we reduce to 45 classes after post-processing (See \textbf{Appendix~\ref{sec:appendix_dataset_full}}) and \textbf{SlideChat-HistAI}~\cite{chen_slidechat_2025}, a diagnostic subtyping task across 3 organs and 1,396 slides. For both slide-level tasks, we formulate this as an organ-specific multiple-choice task, mirroring clinical workflows where tissue origin is known \textit{a priori}. For each WSI, valid diagnostic options are restricted to its corresponding organ. For patch-level evaluation, we use \textbf{SPIDER}~\cite{nechaev_spider_nodate}, a four-organ expert-annotated benchmark with 408,235 annotated patches covering skin, breast, colorectal, and thorax tissue, evaluated in both fine-grained form (SPIDER-F; 13--24 classes per organ) and a coarse unified taxonomy (SPIDER-C). We also evaluate tumor detection on \textbf{CAMELYON-16} using CAMELYON+ annotations~\cite{ehteshami_bejnordi_diagnostic_2017, ling_comprehensive_2025}, yielding 206,800 labeled patches from 48 slides, and out-of-domain cancer detection (2 classes) and tissue classification (11 classes) on \textbf{CATCH}~\cite{wilm_pan-tumor_2022}, comprising 1,251,381 patches from canine cutaneous biopsies.

\textbf{Zero-Shot and Vision-Language Baselines.}
We compare the supervised models against state-of-the-art zero-shot computational pathology models at the patch-level (MI-Zero \cite{lu_visual_2023}, CONCHv1.5\cite{lu_visual-language_2024}, MUSK \cite{xiang_visionlanguage_2025}, QuiltNet \cite{ikezogwo_quilt-1m_nodate}, and PLIP \cite{huang_visuallanguage_2023}), and at the slide-level (TITAN \cite{ding_multimodal_2025}, PRISM \cite{shaikovski_prism_nodate}). All zero-shot models are evaluated following the standard pipeline described in \textbf{Section \ref{subsec:evaluation}}.

\textbf{Supervised Baselines.}
To compare our zero-shot approach against fully supervised baselines, we utilize ABMIL \cite{ilse_attention-based_2018}, CLAM \cite{lu_data-efficient_2021}, DSMIL \cite{li_dual-stream_2021}, TransMIL \cite{shao_transmil_2021}, and WiKG \cite{li_wikg_2024}. Models are trained on the TCGA-OT training split using a standard 45-way cross-entropy objective. During inference, to ensure a fair comparison with our zero-shot organ-specific forumalation, we constrain the supervised predictions by applying an organ-specific mask. The final prediction is designated as the subtype with the highest predicted probability strictly within the slide's known organ category. We follow the standard training recipe for MIL models in pathology \cite{shao_multiple_2025,chen_towards_2024}, further described in \textbf{Section \ref{sec:supervised_implementation}}.
% including a learning rate of $1 \times 10^{-4}$, AdamW optimization with cosine annealing learning rate schedule, and early stopping with 5 epoch patience. Additional details are provided in \textbf{Section \ref{sec:supervised_implementation}}

\textbf{Pretraining and Implementation Details.}
We use $448 \times 448$ pixel patches unless otherwise specified. LiT is conducted for $3$ epochs, a batch size of $1$ and a learning rate of $1 \times 10^{-5}$ with weight decay $1 \times 10^{-5}$ using pre-computed patch-level features from the Trident library\cite{zhang_accelerating_2025}. We use the AdamW optimizer with a cosine annealing learning rate schedule. For pretraining, each training sample is paired with $4$ negative reports for the contrastive loss. A $1\%$ validation split is held out for selecting the best epoch. Text reports were processed with a locally-hosted Qwen3-14B model \cite{qwen3technicalreport} to remove identifying patient information and filter details unrelated to histology (detailed in \textbf{Section \ref{sec:appendix_dataset_full}}). For zero-shot target labels, we follow the CPath convention of prepending the tissue type to the class label~\citep{bahadir_llm-generated_nodate,lu_visual_2023,sharma_investigating_2025}.

\section{Results}

\subsection{Zero-Shot Patch-Level Classification}
We begin our evaluation by investigating the capacity of \ours to perform detailed, multi-class patch-level classification. To validate that our slide-level training regime successfully aligns macroscopic clinical text with fine-grained microscopic morphologies, we evaluate the model on large-scale benchmarks across both human (SPIDER) and canine (CATCH) tissues. Despite being optimized exclusively on slide-level reports, \ours demonstrates highly competitive zero-shot performance across varied tissue types and species.

\begin{table*}[htbp]
\small
\centering
\caption{\textbf{Patch-level zero-shot.} Performance on patch-level tasks for C16 (AUC), SPIDER (balanced accuracy), and CATCH (balanced accuracy). Number of classes shown in parentheses. Best performance in \textbf{bold}; second-best \underline{underlined}. Standard deviation according to 1,000 bootstrap trials.}
\label{tab:patch_zeroshot}
\resizebox{\textwidth}{!}{
\begin{tabular}{llcccccc}
\toprule
\textbf{Dataset} & \textbf{Organ} & \textbf{PLIP\cite{huang_visuallanguage_2023}} & \textbf{QuiltNet\cite{ikezogwo_quilt-1m_nodate}} & \textbf{Musk\cite{xiang_visionlanguage_2025}} & \textbf{Mi-Zero\cite{lu_visual_2023}} & \textbf{CONCH\cite{lu_visual-language_2024}} & \textbf{\ours} \\
\midrule
\multirow{1}{*}{C16 (2)} & Lymph Node (2) & 88.4 $\pm$ 0.1 & \underline{91.7 $\pm$ 0.1} & 82.7 $\pm$ 0.2 & 59.9 $\pm$ 0.2 & 90.3 $\pm$ 0.2 & \textbf{96.7 $\pm$ 0.1} \\
\midrule
\multirow{5}{*}{SPIDER-C} & Skin (9) & 38.0 $\pm$ 0.3 & 30.8 $\pm$ 0.3 & \textbf{62.1 $\pm$ 0.3} & 39.2 $\pm$ 0.3 & 50.8 $\pm$ 0.3 & \underline{60.8 $\pm$ 0.3} \\
 & Colorectal (8) & 43.5 $\pm$ 0.3 & 37.7 $\pm$ 0.4 & 43.5 $\pm$ 0.3 & \underline{47.7 $\pm$ 0.4} & 44.0 $\pm$ 0.4 & \textbf{54.8 $\pm$ 0.4} \\
 & Breast (7) & 32.8 $\pm$ 0.3 & 28.2 $\pm$ 0.5 & 57.8 $\pm$ 0.4 & 50.6 $\pm$ 0.4 & \underline{57.9 $\pm$ 0.4} & \textbf{70.2 $\pm$ 0.3} \\
 & Thorax (7) & 25.0 $\pm$ 0.4 & 23.6 $\pm$ 0.4 & 41.7 $\pm$ 0.4 & 39.2 $\pm$ 0.5 & \underline{46.2 $\pm$ 0.4} & \textbf{49.4 $\pm$ 0.4} \\
\rowcolor{gray!15} & \textbf{Overall} & 34.8 $\pm$ 0.2 & 30.1 $\pm$ 0.2 & \underline{51.3 $\pm$ 0.2} & 44.2 $\pm$ 0.2 & 49.7 $\pm$ 0.2 & \textbf{58.8 $\pm$ 0.2} \\
\midrule
\multirow{5}{*}{SPIDER-F} & Skin (24) & 32.1 $\pm$ 0.2 & 17.4 $\pm$ 0.2 & \underline{50.8 $\pm$ 0.2} & 33.0 $\pm$ 0.2 & 42.3 $\pm$ 0.2 & \textbf{53.3 $\pm$ 0.2} \\
 & Colorectal (13) & 30.9 $\pm$ 0.3 & 28.4 $\pm$ 0.3 & 34.5 $\pm$ 0.3 & 36.7 $\pm$ 0.3 & \underline{38.5 $\pm$ 0.3} & \textbf{48.6 $\pm$ 0.4} \\
 & Breast (18) & 17.3 $\pm$ 0.2 & 11.3 $\pm$ 0.3 & 41.7 $\pm$ 0.4 & 36.5 $\pm$ 0.4 & \underline{46.9 $\pm$ 0.4} & \textbf{52.7 $\pm$ 0.4} \\
 & Thorax (14) & 22.0 $\pm$ 0.2 & 21.5 $\pm$ 0.2 & \underline{49.2 $\pm$ 0.3} & 33.4 $\pm$ 0.3 & 43.8 $\pm$ 0.3 & \textbf{50.7 $\pm$ 0.3} \\
\rowcolor{gray!15} & \textbf{Overall} & 25.6 $\pm$ 0.1 & 19.7 $\pm$ 0.1 & \underline{44.1 $\pm$ 0.2} & 34.9 $\pm$ 0.1 & 42.9 $\pm$ 0.1 & \textbf{51.3 $\pm$ 0.2} \\
\midrule
\multirow{3}{*}{CATCH} & Skin-F (11) & 19.2 $\pm$ 0.0 & 25.3 $\pm$ 0.0 & \underline{44.2 $\pm$ 0.1} & 41.4 $\pm$ 0.1 & 23.4 $\pm$ 0.0 & \textbf{45.1 $\pm$ 0.0} \\
 & Skin-C (2) & \underline{79.4 $\pm$ 0.0} & 71.3 $\pm$ 0.0 & 74.0 $\pm$ 0.0 & 61.2 $\pm$ 0.0 & 72.4 $\pm$ 0.0 & \textbf{83.7 $\pm$ 0.0} \\
\rowcolor{gray!15} & \textbf{Overall} & 49.3 $\pm$ 0.0 & 48.3 $\pm$ 0.0 & \underline{59.1 $\pm$ 0.0} & 51.3 $\pm$ 0.0 & 47.9 $\pm$ 0.0 & \textbf{64.4 $\pm$ 0.0} \\
\bottomrule
\end{tabular}
}
\end{table*}

On the patch-level benchmarks, \ours demonstrates broad improvements over existing vision-language models, establishing new state-of-the-art performance across nearly all evaluated sub-tasks.On the coarse-grained SPIDER-C dataset, \ours establishes new state-of-the-art accuracies in Colorectal (54.8\%), Breast (70.2\%), and Thorax (49.4\%), while remaining competitive on Skin (60.8\%, second to Musk), substantially outperforming models pretrained on patch-caption pairs. A similar trend is observed in the fine-grained SPIDER-F task, where \ours leads across all organ types, achieving top marks in Skin (53.3\%), Colorectal (48.6\%), Breast (52.7\%), and Thorax (50.7\%) to deliver an overall balanced accuracy of 51.3\%. Similarly, we observe consistent improvements on the canine CATCH benchmark, where \ours achieves state-of-the-art performance in both coarse-grained cancer detection (83.7\%) and fine-grained subtyping (45.1\%). Our overall results confirm that locked-image tuning on slide-level reports successfully preserves, and in many cases enhances, microscopic visual-semantic alignment.

\subsection{Zero-Shot Tumor Detection and Spatial Interpretability}
To assess grounding of slide-level text to patch-level visual features without explicit spatial supervision or complex MIL aggregators, we evaluate on the CAMELYON-16 (C16) benchmark which contains pixel-level annotations for breast cancer metastasis detection.
% On c16 metastasis detection, 

\begin{figure}[htbp]
  \centering
  \includegraphics[width=1\linewidth]{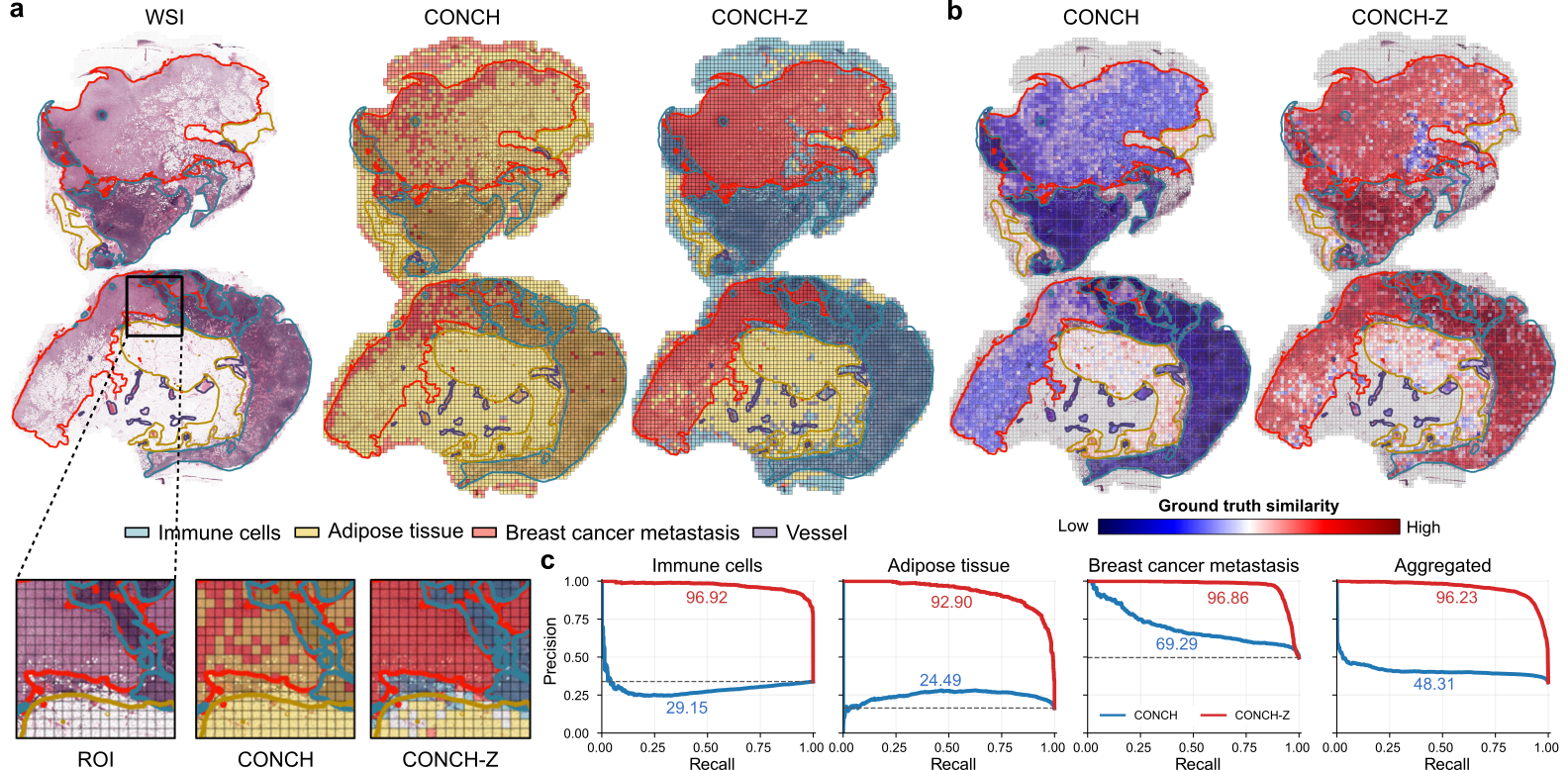}
  \caption{\textbf{Zero-shot breast cancer metastasis detection on the CAMELYON16 dataset}. \textbf{a)} 4-way patch classification performed according to maximum cosine similarity. Metastasis annotations were provided by Camelyon+, while the remaining annotations were generated by an in-house pathologist. \textbf{b)} Model score for the ground truth class. Red indicates the ground truth class received high predicted scores, while blue indicates low scores for the predicted class. Scores are normalized per-patch, and heatmap range is determined per-model. \textbf{c)} Area Under the Precision-Recall Curve for the key tissue categories for CONCHv1.5 (blue) and \ours (red).}
  \label{fig:heatmap}
\end{figure}

\ours achieves an overall zero-shot AUC of 96.7\%, outperforming both QuiltNet (91.7\%) and CONCHv1.5 (90.3\%). Beyond slide-level classification, this fine-grained semantic alignment enables inherent spatial interpretability. As illustrated in \textbf{Figure \ref{fig:heatmap}}, when prompted with key diagnostic terms, the aligned text encoder achieves substantial overlap with pathologist-annotated regions of metastasis, adipose tissue, and lymphocytes. Visualizations for other models are in \textbf{Figure \ref{fig:app_heatmap}}.

\subsection{Zero-Shot Slide-Level Classification}
We next evaluate the representational capacity of our optimized text encoder on slide-level subtyping tasks using the TCGA-OT (n=11,186) and SlideChat-HistAI (n=1,396) datasets. \ours establishes a new state-of-the-art across both benchmarks, outperforming heavily parameterized slide-level foundation models as well as established patch-level baselines (\textbf{Table \ref{tab:slide_zeroshot}}).

On the TCGA-OT benchmark, \ours achieves an overall zero-shot balanced accuracy of 78.8\%, representing an absolute improvement of +21.4\% over its base vision backbone (CONCH, 57.4\%) and a +5.7\% improvement over the leading slide-level foundation model, TITAN (73.1\%, compared to PRISM at 59.4\%). We visualize this stronger alignment of the text and visual features in \textbf{Fig.~\ref{fig:app_tsne}}.

This performance gap persists on the HistAI dataset, which poses a 3-class classification task for breast, colorectum, and skin cancers. \ours achieves an overall balanced accuracy of 54.6\%, outperforming the next best patch model (Musk, 44.6\%) and slide model (PRISM, 50.2\%) by a wide margin. These results underscore that fine-tuning the text modality on clinical reports bridges the semantic gap required for complex, organ-specific diagnostic reasoning.

\begin{table*}[htbp]
\tiny
\centering
\caption{\textbf{Slide-level zero-shot.} HistAI and TCGA-OT per-organ balanced accuracy. Best \textbf{bold}; second-best \underline{underlined}. Standard deviation reported according to 1,000 bootstrap trials.}
\label{tab:slide_zeroshot}
\resizebox{\textwidth}{!}{
\begin{tabular}{llcccccc|cc}
\toprule
\multirow{2}{*}{Dataset} & \multirow{2}{*}{Organ} & \multicolumn{6}{c}{\textbf{Patch foundation model}} & \multicolumn{2}{c}{\textbf{Slide foundation model}} \\
\cmidrule(lr){3-8} \cmidrule(lr){9-10}
 &  & PLIP\cite{huang_visuallanguage_2023} & QuiltNet\cite{ikezogwo_quilt-1m_nodate} & Musk\cite{xiang_visionlanguage_2025} & Mi-Zero\cite{lu_visual_2023} & CONCH\cite{lu_visual-language_2024} & \ours & TITAN\cite{ding_multimodal_2025} & PRISM\cite{shaikovski_prism_nodate} \\
\midrule
\multirow{4}{*}{HistAI} & Breast & 41.6 $\pm$ 3.6 & 32.3 $\pm$ 1.3 & \underline{49.5 $\pm$ 3.6} & 44.7 $\pm$ 2.8 & 34.0 $\pm$ 1.4 & 47.6 $\pm$ 3.6 & 46.1 $\pm$ 3.5 & \textbf{49.7 $\pm$ 2.7} \\
 & Colorectum & 28.4 $\pm$ 2.2 & 33.8 $\pm$ 3.6 & 36.2 $\pm$ 3.6 & 31.4 $\pm$ 2.9 & 32.9 $\pm$ 3.1 & \textbf{51.4 $\pm$ 4.2} & 33.2 $\pm$ 2.9 & \underline{44.3 $\pm$ 3.8} \\
 & Skin & 47.6 $\pm$ 3.1 & 33.3 $\pm$ 0.0 & 48.1 $\pm$ 3.3 & 46.0 $\pm$ 3.1 & 53.1 $\pm$ 3.7 & \textbf{65.0 $\pm$ 3.1} & \underline{58.9 $\pm$ 3.0} & 56.7 $\pm$ 3.7 \\
\rowcolor{gray!15} & \textbf{Overall} & 39.2 $\pm$ 1.7 & 33.1 $\pm$ 1.3 & 44.6 $\pm$ 2.1 & 40.7 $\pm$ 1.7 & 40.0 $\pm$ 1.8 & \textbf{54.6 $\pm$ 2.2} & 46.1 $\pm$ 1.7 & \underline{50.2 $\pm$ 2.0} \\
\midrule
\multirow{9}{*}{TCGA} & Breast & 48.8 $\pm$ 0.6 & 7.2 $\pm$ 0.9 & 62.8 $\pm$ 1.6 & 49.6 $\pm$ 1.8 & 50.0 $\pm$ 0.6 & \textbf{89.2 $\pm$ 1.2} & 77.7 $\pm$ 1.7 & \underline{81.9 $\pm$ 1.3} \\
 & Lung & 52.5 $\pm$ 3.1 & 27.8 $\pm$ 4.4 & \underline{68.1 $\pm$ 3.2} & 31.9 $\pm$ 4.2 & 47.4 $\pm$ 3.9 & \underline{74.2 $\pm$ 4.0} & \textbf{95.2 $\pm$ 0.9} & 57.1 $\pm$ 3.8 \\
 & Brain & 17.9 $\pm$ 1.2 & 33.4 $\pm$ 6.5 & 15.6 $\pm$ 1.5 & 14.7 $\pm$ 1.6 & \underline{39.2 $\pm$ 5.0} & \textbf{44.4 $\pm$ 4.6} & 33.0 $\pm$ 5.5 & 26.9 $\pm$ 6.0 \\
 & Kidney & 66.9 $\pm$ 1.8 & 58.4 $\pm$ 1.9 & 57.2 $\pm$ 1.5 & 75.5 $\pm$ 1.7 & 77.9 $\pm$ 2.6 & \textbf{96.4 $\pm$ 0.7} & \underline{83.0 $\pm$ 2.0} & 67.0 $\pm$ 3.8 \\
 & Thyroid & 25.4 $\pm$ 1.9 & 45.2 $\pm$ 2.6 & 52.3 $\pm$ 2.6 & 67.7 $\pm$ 2.2 & \underline{76.2 $\pm$ 2.4} & \textbf{82.9 $\pm$ 2.1} & 74.6 $\pm$ 1.8 & 62.9 $\pm$ 2.6 \\
 & Bowel & 39.2 $\pm$ 5.4 & 40.3 $\pm$ 3.1 & 57.5 $\pm$ 4.3 & 39.6 $\pm$ 7.4 & 58.2 $\pm$ 4.7 & \textbf{74.7 $\pm$ 2.1} & \underline{64.7 $\pm$ 2.5} & 51.6 $\pm$ 4.8 \\
 & Uterus & 34.0 $\pm$ 2.7 & 50.2 $\pm$ 0.2 & 45.1 $\pm$ 2.6 & 35.4 $\pm$ 2.3 & 46.8 $\pm$ 1.2 & \textbf{82.1 $\pm$ 2.1} & \underline{80.0 $\pm$ 1.5} & 65.4 $\pm$ 2.6 \\
 & Other & 38.2 $\pm$ 3.3 & 35.4 $\pm$ 3.4 & 49.5 $\pm$ 3.5 & 37.6 $\pm$ 3.4 & 63.3 $\pm$ 3.4 & \textbf{86.7 $\pm$ 2.8} & \underline{76.3 $\pm$ 2.6} & 62.4 $\pm$ 3.2 \\
\rowcolor{gray!15} & \textbf{Overall} & 40.4 $\pm$ 1.6 & 37.2 $\pm$ 1.4 & 51.0 $\pm$ 1.4 & 44.0 $\pm$ 1.5 & 57.4 $\pm$ 1.7 & \textbf{78.8 $\pm$ 1.7} & \underline{73.1 $\pm$ 0.9} & 59.4 $\pm$ 1.7 \\
\bottomrule
\end{tabular}
}
\end{table*}

\subsection{Comparison with Supervised Baselines and Learned Aggregation}
We next sought to contextualize the performance of zero-shot classification over popular supervised approaches. We evaluate \ours against a suite of fully supervised MIL architectures on the TCGA-OT test set (1,348 slides). The supervised models (ABMIL, TransMIL, DSMIL, WiKG, and CLAM) were trained for 45-way classification on 8,226 samples from the official TCGA-OT splits, with melanoma and cutaneous melanoma grouped into a single class due to label noise (See \textbf{Appendix ~\ref{sec:appendix_dataset_full}}). To ensure a fair comparison with our zero-shot formulation, which performed inference on disease subtypes within each organ category, we select the highest logit within-organ prediction during the evaluation of these supervised models as well.

\begin{table*}[htbp]
\tiny
\centering
\setlength{\tabcolsep}{3.5pt}
\caption{
            \textbf{{TCGA-OT supervised baselines}} Supervised baselines and linear probes were trained on 8,226 samples and evaluated on 1,348 samples. Balanced accuracy is reported for all cases ($\pm$ standard deviation). \textbf{Overall} represents the average balanced accuracy across all organs.
            }
\label{tab:tcga_OT_supervised_patch_test}
\resizebox{\textwidth}{!}{
\begin{tabular}{llcccccccc|c}
\toprule
Type & Model & \textbf{Breast} & \textbf{Lung} & \textbf{Brain} & \textbf{Kidney} & \textbf{Thyroid} & \textbf{Bowel} & \textbf{Uterus} & \textbf{Other} & \textbf{Overall} \\
\midrule
\parbox[t]{0mm}{\multirow{7}{*}{\rotatebox[origin=c]{90}{\textbf{Supervised}}}} & ABMIL\cite{ilse_attention-based_2018} & \textbf{91.0 $\pm$ 3.6} & \underline{88.5 $\pm$ 2.6} & 26.5 $\pm$ 4.5 & \underline{89.4 $\pm$ 3.2} & 57.3 $\pm$ 3.0 & \textbf{63.1 $\pm$ 10.1} & 81.8 $\pm$ 5.1 & \underline{89.8 $\pm$ 1.0} & 73.4 $\pm$ 0.9 \\
 & TransMIL\cite{shao_transmil_2021} & 63.6 $\pm$ 5.5 & \textbf{90.8 $\pm$ 2.4} & 31.8 $\pm$ 4.8 & 87.7 $\pm$ 3.5 & \underline{68.1 $\pm$ 3.4} & 58.6 $\pm$ 10.6 & 78.6 $\pm$ 5.5 & 88.8 $\pm$ 1.2 & 71.0 $\pm$ 1.1 \\
 & DSMIL\cite{li_dual-stream_2021} & 80.9 $\pm$ 5.3 & 87.9 $\pm$ 3.0 & 32.8 $\pm$ 5.5 & 88.7 $\pm$ 3.5 & 48.8 $\pm$ 1.3 & 59.6 $\pm$ 10.7 & \textbf{84.6 $\pm$ 4.3} & 85.5 $\pm$ 2.7 & 71.1 $\pm$ 2.0 \\
 & CLAM\cite{lu_data-efficient_2021} & \underline{88.9 $\pm$ 4.0} & 89.7 $\pm$ 2.6 & \textbf{36.8 $\pm$ 5.8} & 82.0 $\pm$ 4.0 & 50.2 $\pm$ 1.5 & 52.2 $\pm$ 10.4 & \underline{84.1 $\pm$ 4.2} & 88.2 $\pm$ 1.3 & 71.5 $\pm$ 1.2 \\
 & WiKG\cite{li_wikg_2024} & 88.5 $\pm$ 4.6 & 88.0 $\pm$ 2.9 & \underline{36.6 $\pm$ 5.9} & 88.2 $\pm$ 3.7 & 47.5 $\pm$ 0.4 & 56.9 $\pm$ 10.4 & 73.1 $\pm$ 5.7 & 88.0 $\pm$ 1.2 & 70.9 $\pm$ 1.1 \\
 & $\text{CONCH}_{\text{lin}}$\cite{lu_visual-language_2024} & 83.9 $\pm$ 5.2 & 87.1 $\pm$ 3.1 & 30.1 $\pm$ 7.5 & 89.7 $\pm$ 3.3 & \textbf{80.4 $\pm$ 3.2} & \underline{62.9 $\pm$ 9.2} & 79.2 $\pm$ 5.5 & \underline{88.9 $\pm$ 1.1} & \underline{75.3 $\pm$ 1.1} \\
 \rowcolor{gray!15} & $\text{CONCH-Z}_{\text{lin}}$ & 83.9 $\pm$ 5.2 & 86.8 $\pm$ 3.2 & 33.4 $\pm$ 7.9 & \textbf{90.7 $\pm$ 3.1} & \textbf{80.4 $\pm$ 3.2} & \underline{62.9 $\pm$ 9.2} & 78.9 $\pm$ 5.5 & \textbf{90.8 $\pm$ 1.1} & \textbf{76.0 $\pm$ 1.1} \\
\midrule
\parbox[t]{0mm}{\multirow{8}{*}{\rotatebox[origin=c]{90}{\textbf{Zero-shot}}}} & TITAN\cite{ding_multimodal_2025} & \underline{83.5 $\pm$ 5.1} & \underline{91.4 $\pm$ 2.5} & 32.2 $\pm$ 6.4 & 77.1 $\pm$ 4.2 & 83.7 $\pm$ 2.5 & 66.0 $\pm$ 9.3 & \underline{83.2 $\pm$ 3.5} & 61.7 $\pm$ 1.8 & \underline{72.4 $\pm$ 1.4} \\
 & PRISM\cite{shaikovski_prism_nodate} & 82.0 $\pm$ 4.9 & 60.6 $\pm$ 9.7 & 20.4 $\pm$ 5.9 & 58.4 $\pm$ 6.3 & 66.5 $\pm$ 4.0 & 65.6 $\pm$ 9.1 & 62.6 $\pm$ 6.0 & \underline{74.2 $\pm$ 1.9} & 61.3 $\pm$ 1.6 \\
 & MI-Zero\cite{lu_visual_2023} & 52.6 $\pm$ 5.0 & 45.8 $\pm$ 13.2 & 19.2 $\pm$ 5.9 & 73.4 $\pm$ 5.2 & 73.4 $\pm$ 3.2 & 47.4 $\pm$ 10.9 & 48.9 $\pm$ 4.0 & 40.9 $\pm$ 3.8 & 50.2 $\pm$ 2.6 \\
 & PLIP\cite{huang_visuallanguage_2023} & 50.0 $\pm$ 0.0 & 75.0 $\pm$ 6.3 & 12.5 $\pm$ 4.7 & 65.4 $\pm$ 5.2 & 27.7 $\pm$ 4.2 & 51.0 $\pm$ 10.7 & 41.9 $\pm$ 8.6 & 42.9 $\pm$ 4.0 & 45.8 $\pm$ 2.7 \\
 & QuiltNet\cite{ikezogwo_quilt-1m_nodate} & 19.6 $\pm$ 5.9 & 46.0 $\pm$ 13.1 & 20.0 $\pm$ 6.9 & 57.5 $\pm$ 5.7 & 52.0 $\pm$ 3.8 & 58.5 $\pm$ 10.2 & 51.4 $\pm$ 5.7 & 26.0 $\pm$ 2.7 & 41.4 $\pm$ 2.1 \\
 & Musk\cite{xiang_visionlanguage_2025} & 73.5 $\pm$ 5.0 & 69.0 $\pm$ 7.7 & 12.5 $\pm$ 4.5 & 58.0 $\pm$ 4.8 & 62.6 $\pm$ 3.9 & \textbf{67.4 $\pm$ 8.7} & 52.4 $\pm$ 6.2 & 57.3 $\pm$ 3.8 & 56.6 $\pm$ 2.7 \\
 & CONCH\cite{lu_visual-language_2024} & 49.8 $\pm$ 2.8 & 80.3 $\pm$ 4.9 & \textbf{62.4 $\pm$ 5.1} & \underline{79.0 $\pm$ 4.5} & \underline{85.7 $\pm$ 2.9} & 54.7 $\pm$ 8.0 & 50.0 $\pm$ 0.0 & 70.0 $\pm$ 3.5 & 66.5 $\pm$ 2.4 \\
 \rowcolor{gray!15} & \ours & \textbf{90.7 $\pm$ 4.2} & \textbf{91.5 $\pm$ 2.6} & \underline{33.7 $\pm$ 6.8} & \textbf{89.4 $\pm$ 3.6} & \textbf{92.4 $\pm$ 2.1} & \underline{66.1 $\pm$ 9.6} & \textbf{83.2 $\pm$ 4.8} & \textbf{92.8 $\pm$ 1.1} & \textbf{80.0 $\pm$ 1.0} \\
\bottomrule
\end{tabular}
}
\end{table*}

Furthermore, we directly isolate and evaluate the efficacy of our zero-shot models against learned aggregation schemes with an additional linear probe baseline. In this setup, we utilize the base patch encoders (CONCH and \ours) paired with the text query ``cancer'' to serve as a zero-shot spatial aggregator. Subsequently, a linear probe (\conchlin and \ourslin) is trained on these aggregated representations to produce the final prediction. Remarkably, this fixed, zero-shot aggregation performs competitively with supervised aggregators, with \ourslin achieving 76.0\% overall balanced accuracy, outperforming the best-performing traditional MIL model (ABMIL, 73.4\%) by $+2.6\%$ and trailing only behind its supervised counterpart. Meanwhile, using \ours for end-to-end zero-shot classification achieves an overall balanced accuracy of 80.0\% (\textbf{Table~\ref{tab:tcga_OT_supervised_patch_test}}), representing a gain of $+6.6\%$ over the best supervised MIL approach (ABMIL, 73.4\%) and $+4.0\%$ over the fixed-query linear probe, \ourslin (76.0\%). These findings highlight the importance of query-aware aggregation paired with zero-shot classification.

\subsection{Ablations}

\begin{figure}[t]
    \centering
    % Left side: Figure
    \begin{minipage}[c]{0.48\textwidth}
        \centering
        \includegraphics[width=\textwidth]{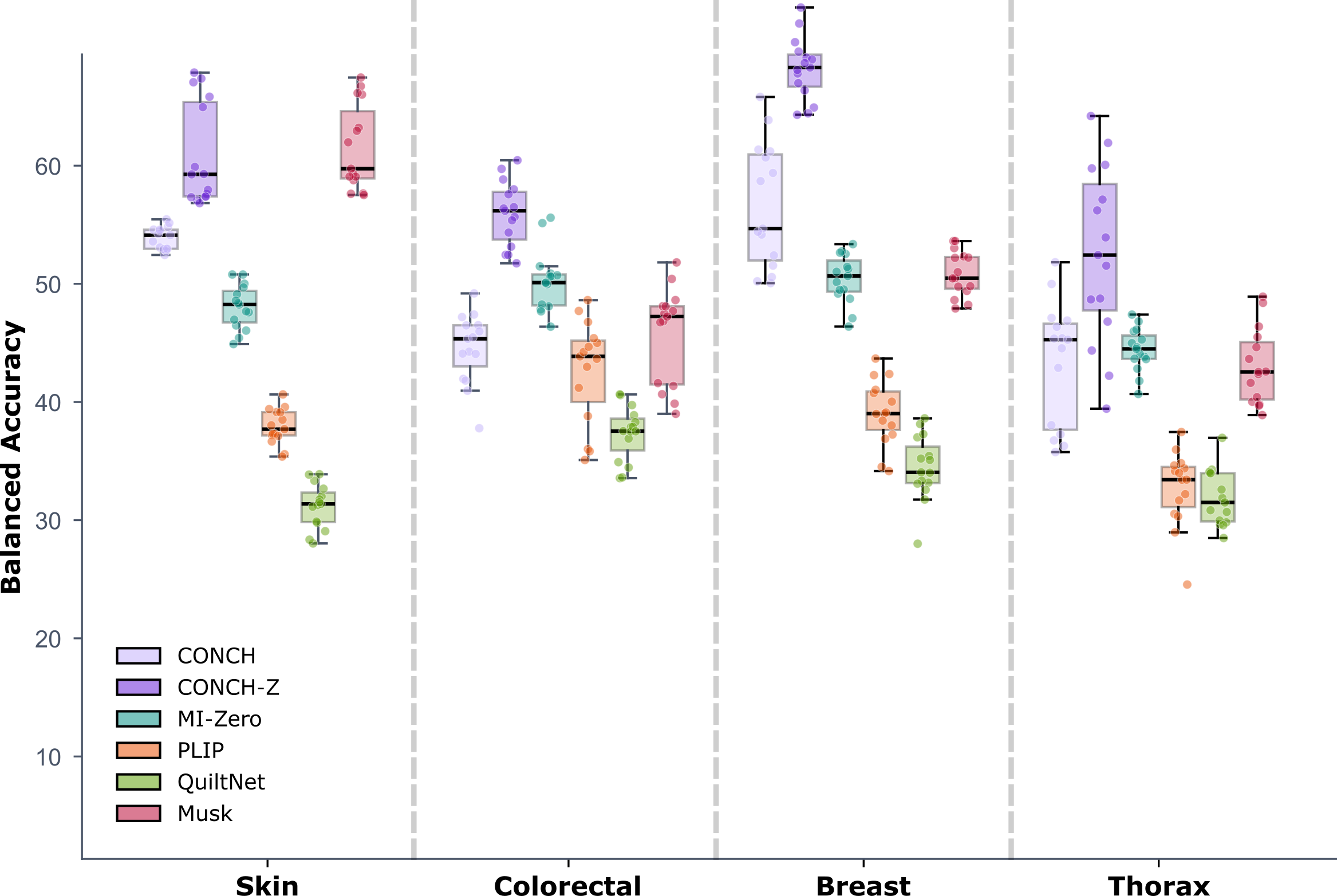}
        \captionof{figure}{SPIDER-C tissue subtyping performance when the cancer class prompt is substituted with every combination of five cancer-related synonyms and three sentence templates. Each point represents performance with a single synonym-template pair.}
        \label{fig:ablation_spider}
    \end{minipage}\hfill
    \begin{minipage}[c]{0.48\textwidth}
        \centering
        \tiny
        \captionof{table}{Ablation studies. Effect of various design choices for training and inference on balanced accuracy, averaged across TCGA-OT (n=11,186) and HistAI (n=1,396)}
       \label{tab:ablation_design}
\begin{tabular}{cllccc}
    \toprule
    & & & \multicolumn{3}{c}{\textbf{Pretrain Size}} \\
    \cmidrule(lr){4-6}
    & \textbf{Ablation} & \textbf{Setting} & \textbf{340k} & \textbf{150k} & \textbf{50k} \\
    \midrule
    \multirow{9}{*}{\rotatebox[origin=c]{90}{\scriptsize\textbf{Training}}} 
    & \textbf{Full model} & Ours & 66.6\% & 64.8\% & 62.7\% \\
    \cmidrule{2-6}
    & \multirow{5}{*}{\parbox[l]{1.0cm}{Negative\\selection}}
    & Rand. organ & 64.7\% & 64.0\% & 60.5\% \\
    & & $N=8$      & 65.7\% & 62.0\% & 62.1\% \\
    & & $N=6$      & 65.4\% & 64.8\% & 63.0\% \\
    & & $N=2$      & 63.9\% & 62.6\% & 61.7\% \\
    & & $N=1$      & 61.9\% & 62.0\% & 60.5\% \\
    \cmidrule{2-6}
    & \multirow{2}{*}{\parbox[l]{1.0cm}{Loss}}
    & No weight & 58.5\% & 59.1\% & 59.1\% \\
    & & SigLIT \cite{zhai_sigmoid_2023}    & 66.3\%     & 65.1\%     & 61.5\%     \\
    \cmidrule{2-6}
    & Finetune & LoRA\cite{hu_lora_2021} & 64.1\% & 63.7\% & 61.8\% \\
    \midrule
    \multirow{4}{*}{\rotatebox[origin=c]{90}{\scriptsize\textbf{Inference}}}
    & \multirow{4}{*}{\parbox[l]{1cm}{Pooling method}} 
    & Mean pool & 61.5\% & 59.5\% & 60.6\% \\
    & & Rand. pool & 59.0\% & 57.5\% & 55.7\% \\
    & & Top-10 & 65.2\% & 64.6\% & 60.4\% \\
    & & Top-100 & 66.7\% & 63.0\% & 62.7\% \\
    \bottomrule
\end{tabular}
    \end{minipage}
\end{figure}

We evaluate the influence of various training-time and inference-time decisions to assess the robustness of our approach in \textbf{Table \ref{tab:ablation_design}}, which reports the overall accuracy averegad across TCGA-OT and HistAI tasks after altering various train-time and test-time approaches. 

\textbf{Pretrain Size}: We consider three sizes of the pretraining set: the full \nslides samples, 150k samples, and 50k samples, in order to provide greater insight into how each design decision influences performance at various pretrain scales. Following LIMIT pretraining, we observe steady performance scaling with data volume across all evaluation configurations, with balanced accuracy increasing consistently from 62.7\% at 50k to 66.6\% at 340k for our primary model.

\textbf{Negative selection}: We investigate various approaches for selecting $N$ negative reports during training. While our standard training approach only retrieves reports from the same organ, allowing random organ selection causes a 2.9\% relative drop at 340k and a 3.5\% drop at 50k (1.9 and 2.2 percentage points, respectively). We also vary the number of negative reports retrieved from 1 to 8. While \ours samples 4 negative reports per batch, performance remains robust when sampling larger numbers of negatives ($N=6$ or $8$) with minimal degradation (up to a 1.8\% relative drop at 340k), but degrades more noticeably when using fewer negatives ($N=2$ or $1$, falling by up to 7.1\% relatively at 340k), indicating that a sufficient pool of hard in-organ negatives is essential for effective contrastive alignment.

\textbf{Loss}: We remove the weighted loss scaling, such that all reports are equally weighted regardless of embedding similarity to the positive pair report. Removing this weighted loss causes a substantial drop across all regimes, decreasing performance by 8.1 percentage points (a 12.2\% relative drop) at 340k, which confirms that weight scaling by report similarity is critical to prevent penalizing semantically congruent reports. Replacing the InfoNCE loss with weighted SigLIT loss \cite{zhai_sigmoid_2023}, a prominent contrastive loss for LiT, yields performance closely tracking InfoNCE across all pretraining regimes (e.g., 66.3\% versus 66.6\% at 340k), suggesting that LIMIT pretraining is robust to the choice of contrastive objective.

\textbf{Finetune}: Rather than finetuning the full text encoder weights, we finetune with low-rank adaptation (LoRA) \cite{hu_lora_2021}. Although LoRA provides a parameter-efficient alternative, performance gains remain lower across all dataset scales (e.g., 64.1\% versus 66.6\% at 340k), indicating that updating the full set of text representation weights is advantageous for fine-grained clinical domain transfer.

\textbf{Pooling method}: Rather than pool \ours with weighted similarity to each text option at inference-time, we perform mean pooling and random pooling (reported as the average of ten trials per task). Replacing LIMIT pooling with mean or random pooling leads to noticeable relative drops of 3--8\% and 11\%, highlighting the importance of content-aware aggregation. Aggregating the slide from top-10 and top-100 patches yields comparable performance to weighted aggregation of all patches with LIMIT, but results vary across training scales. We hypothesize that LIMIT's aggregation of all patches may be more robust to heterogeneous tissue distributions than top-$k$ aggregation, which relies on appropriate selection of $k$.  
See \textbf{Appendix \ref{sec:ablation_losses}} for further details on ablation implementation.

\textbf{Synonym evaluation}: We next evaluate the robustness of performance to variations in terminology and phrasing by constructing four additional terms with similar meaning to cancer (tumor, malignant neoplasm, carcinoma, malignancy) and three distinct sentence templates, described in \textbf{Table \ref{tab:spider_terms_templates}}.

The results demonstrate that our approach maintains strong, consistent performance across different semantic and syntactic structures. Regardless of the specific synonym or template combination utilized, our model robustly outperforms existing state-of-the-art vision-language baselines, including the frozen CONCHv1.5 backbone. By achieving a higher median balanced accuracy across all text variations, these findings indicate that the representational alignment achieved through our training strategy yields intrinsic, prompt-agnostic improvements for zero-shot classification.

\textbf{Limitations:} A distinguishing feature of LIMIT is that it introduces zero tunable parameters on the image side: no slide encoder, learned aggregator or visual projection head, with all adaptation flowing through the text encoder via report-conditioned attention. This design preserves the patch-level visual space and reduces training cost, but may be less expressive for holistic slide-level properties such as grade, tumor cellularity, extent of necrosis, or global mitotic activity, which may require modeling relationships across many regions. Our current evaluation focuses primarily on diagnosis, subtyping, and tissue classification, so future work should further assess such global slide-level tasks and explore lightweight image-side adaptation while retaining the benefits of locked-image tuning.
% A distinguishing feature of LIMIT is that it introduces zero tunable parameters on the image side: no slide encoder, learned aggregator or visual projection head, with all adaptation flowing through the text encoder via report-conditioned attention. While this design yields a remarkably parameter-efficient training regime, it likely limits the model's ability to capture holistic, slide-spanning descriptors (e.g. overall grade, extent of necrosis, percent tumor cellularity, and global mitotic activity) that are sometimes reported by pathologists but may be difficult to recover from cosine-weighted aggregation over individual ROIs; introducing a small learnable image-side module (e.g., a 1–2 layer MLP or a single transformer block) is a natural next step. Relatedly, our 20-task evaluation currently centers on diagnosis, subtyping, and tissue classification, and does not directly probe the holistic tasks (grading, mitotic counting, percent involvement) where this architectural limitation would most likely manifest. Finally, by construction, \ours cannot recover morphological features that the frozen CONCH v1.5 backbone fails to encode, underscoring that continued progress on patch-level vision encoders remains complementary to text-side alignment.

\section{Conclusion}
\vspace{-1mm}
We introduce LIMIT, a locked-image tuning framework that aligns patch-level pathology vision-language models with slide-level clinical reports through report-conditioned attention. Using LIMIT, \ours improves zero-shot patch- and slide-level classification across diverse pathology benchmarks while avoiding the parameter and compute overhead of training a dedicated slide encoder. Overall, these results suggest that routine diagnostic reports provide a scalable and underused source of supervision for strengthening visual-semantic alignment in computational pathology.

%\newpage
%\section*{References}
\bibliography{references,references_custom}

@article{bahadir_llm-generated_nodate,
	title = {{LLM}-{Generated} {Rewrite} and {Context} {Modulation} for {Enhanced} {Vision} {Language} {Models} in {Digital} {Pathology}},
	language = {en},
	author = {Bahadir, Cagla Deniz and Akar, Gozde B and Sabuncu, Mert R},
}

@misc{sharma_investigating_2025,
	title = {Investigating {Zero}-{Shot} {Diagnostic} {Pathology} in {Vision}-{Language} {Models} with {Efficient} {Prompt} {Design}},
	url = {http://arxiv.org/abs/2505.00134},
	doi = {10.48550/arXiv.2505.00134},
	urldate = {2026-10-08},
	publisher = {arXiv},
	author = {Sharma, Vasudev and Alagha, Ahmed and Khellaf, Abdelhakim and Trinh, Vincent Quoc-Huy and Hosseini, Mahdi S.},
	month = apr,
	year = {2025},
	note = {arXiv:2505.00134 [cs.CV]},
}

@misc{lu_visual_2023,
	title = {Visual {Language} {Pretrained} {Multiple} {Instance} {Zero}-{Shot} {Transfer} for {Histopathology} {Images}},
	url = {http://arxiv.org/abs/2306.07831},
	doi = {10.48550/arXiv.2306.07831},
	urldate = {2026-09-23},
	publisher = {arXiv},
	author = {Lu, Ming Y. and Chen, Bowen and Zhang, Andrew and Williamson, Drew F. K. and Chen, Richard J. and Ding, Tong and Le, Long Phi and Chuang, Yung-Sung and Mahmood, Faisal},
	month = jun,
	year = {2023},
	note = {arXiv:2306.07831 [cs.CV]},
}

@misc{zhai_sigmoid_2023,
	title = {Sigmoid {Loss} for {Language} {Image} {Pre}-{Training}},
	url = {http://arxiv.org/abs/2303.15343},
	doi = {10.48550/arXiv.2303.15343},
	urldate = {2026-05-06},
	publisher = {arXiv},
	author = {Zhai, Xiaohua and Mustafa, Basil and Kolesnikov, Alexander and Beyer, Lucas},
	month = sep,
	year = {2023},
	note = {arXiv:2303.15343 [cs]},
}

@misc{zhang_accelerating_2025,
	title = {Accelerating {Data} {Processing} and {Benchmarking} of {AI} {Models} for {Pathology}},
	url = {http://arxiv.org/abs/2502.06750},
	doi = {10.48550/arXiv.2502.06750},
	urldate = {2026-05-01},
	publisher = {arXiv},
	author = {Zhang, Andrew and Jaume, Guillaume and Vaidya, Anurag and Ding, Tong and Mahmood, Faisal},
	month = feb,
	year = {2025},
	note = {arXiv:2502.06750 [cs]},
}

@misc{ahmed_pathalign_2024,
	title = {{PathAlign}: {A} vision-language model for whole slide images in histopathology},
	shorttitle = {{PathAlign}},
	url = {http://arxiv.org/abs/2406.19578},
	doi = {10.48550/arXiv.2406.19578},
	urldate = {2026-05-01},
	publisher = {arXiv},
	author = {Ahmed, Faruk and Sellergren, Andrew and Yang, Lin and Xu, Shawn and Babenko, Boris and Ward, Abbi and Olson, Niels and Mohtashamian, Arash and Matias, Yossi and Corrado, Greg S. and Duong, Quang and Webster, Dale R. and Shetty, Shravya and Golden, Daniel and Liu, Yun and Steiner, David F. and Wulczyn, Ellery},
	month = jun,
	year = {2024},
	note = {arXiv:2406.19578 [cs]},
}

@article{wilm_pan-tumor_2022,
	title = {Pan-tumor {CAnine} {cuTaneous} {Cancer} {Histology} ({CATCH}) dataset},
	volume = {9},
	copyright = {2022 The Author(s)},
	issn = {2052-4463},
	url = {https://www.nature.com/articles/s41597-022-01692-w},
	doi = {10.1038/s41597-022-01692-w},
	language = {en},
	number = {1},
	urldate = {2026-05-01},
	journal = {Scientific Data},
	publisher = {Nature Publishing Group},
	author = {Wilm, Frauke and Fragoso, Marco and Marzahl, Christian and Qiu, Jingna and Puget, Chloé and Diehl, Laura and Bertram, Christof A. and Klopfleisch, Robert and Maier, Andreas and Breininger, Katharina and Aubreville, Marc},
	month = sep,
	year = {2022},
	pages = {588},
}

@misc{nechaev_histai_2025,
	title = {{HISTAI}: {An} {Open}-{Source}, {Large}-{Scale} {Whole} {Slide} {Image} {Dataset} for {Computational} {Pathology}},
	shorttitle = {{HISTAI}},
	url = {http://arxiv.org/abs/2505.12120},
	doi = {10.48550/arXiv.2505.12120},
	urldate = {2026-05-01},
	publisher = {arXiv},
	author = {Nechaev, Dmitry and Pchelnikov, Alexey and Ivanova, Ekaterina},
	month = may,
	year = {2025},
	note = {arXiv:2505.12120 [eess]},
}

@article{ling_comprehensive_2025,
	title = {Comprehensive {Benchmark} {Dataset} for {Pathological} {Lymph} {Node} {Metastasis} in {Breast} {Cancer} {Sections}},
	volume = {12},
	issn = {2052-4463},
	doi = {10.1038/s41597-025-05586-5},
	language = {eng},
	number = {1},
	journal = {Scientific Data},
	author = {Ling, Xitong and Lei, Yuanyuan and Li, Jiawen and Cheng, Junru and Huang, Wenting and Guan, Tian and Guan, Jian and He, Yonghong},
	month = aug,
	year = {2025},
	pages = {1381},
}

@article{nechaev_spider_nodate,
	title = {{SPIDER}: {A} {COMPREHENSIVE} {MULTI}-{ORGAN} {SUPERVISED} {PATHOLOGY} {DATASET} {AND} {BASELINE} {MODELS}},
	language = {en},
	author = {Nechaev, Dmitry and Pchelnikov, Alexey and Ivanova, Ekaterina},
}

@article{ikezogwo_quilt-1m_nodate,
	title = {Quilt-{1M}: {One} {Million} {Image}-{Text} {Pairs} for {Histopathology}},
	language = {en},
	author = {Ikezogwo, Wisdom O and Seyfioglu, Mehmet S and Ghezloo, Fatemeh and Geva, Dylan and Mohammed, Fatwir S and Anand, Pavan K and Krishna, Ranjay and Shapiro, Linda G},
}

@article{huang_visuallanguage_2023,
	title = {A visual–language foundation model for pathology image analysis using medical {Twitter}},
	volume = {29},
	copyright = {2023 The Author(s), under exclusive licence to Springer Nature America, Inc.},
	issn = {1546-170X},
	url = {https://www.nature.com/articles/s41591-023-02504-3},
	doi = {10.1038/s41591-023-02504-3},
	language = {en},
	number = {9},
	urldate = {2026-04-15},
	journal = {Nature Medicine},
	publisher = {Nature Publishing Group},
	author = {Huang, Zhi and Bianchi, Federico and Yuksekgonul, Mert and Montine, Thomas J. and Zou, James},
	month = sep,
	year = {2023},
	pages = {2307--2316},
}

@article{shaikovski_prism_nodate,
	title = {{PRISM}: {A} {MULTI}-{MODAL} {GENERATIVE} {FOUNDATION} {MODEL} {FOR} {SLIDE}-{LEVEL} {HISTOPATHOLOGY}},
	language = {en},
	author = {Shaikovski, George and Casson, Adam and Severson, Kristen and Zimmermann, Eric and Wang, Yi Kan and Kunz, Jeremy D and Retamero, Juan A and Oakley, Gerard and Klimstra, David and Kanan, Christopher and Hanna, Matthew and Viret, Julian and Tenenholtz, Neil and Hall, James and Fusi, Nicolo and Yousfi, Razik and Hamilton, Peter and Moye, A and Vorontsov, Eugene and Liu, Siqi and Fuchs, Thomas J},
}

@misc{liang_wsi-llava_2025,
	title = {{WSI}-{LLaVA}: {A} {Multimodal} {Large} {Language} {Model} for {Whole} {Slide} {Image}},
	shorttitle = {{WSI}-{LLaVA}},
	url = {http://arxiv.org/abs/2412.02141},
	doi = {10.48550/arXiv.2412.02141},
	urldate = {2026-04-06},
	publisher = {arXiv},
	author = {Liang, Yuci and Lyu, Xinheng and Chen, Wenting and Ding, Meidan and Zhang, Jipeng and He, Xiangjian and Wu, Song and Xing, Xiaohan and Yang, Sen and Wang, Xiyue and Shen, Linlin},
	month = aug,
	year = {2025},
	note = {arXiv:2412.02141 [cs]},
}

@misc{chen_slidechat_2025,
	title = {{SlideChat}: {A} {Large} {Vision}-{Language} {Assistant} for {Whole}-{Slide} {Pathology} {Image} {Understanding}},
	shorttitle = {{SlideChat}},
	url = {http://arxiv.org/abs/2410.11761},
	doi = {10.48550/arXiv.2410.11761},
	urldate = {2026-04-06},
	publisher = {arXiv},
	author = {Chen, Ying and Wang, Guoan and Ji, Yuanfeng and Li, Yanjun and Ye, Jin and Li, Tianbin and Hu, Ming and Yu, Rongshan and Qiao, Yu and He, Junjun},
	month = mar,
	year = {2025},
	note = {arXiv:2410.11761 [cs]
version: 3},
}

@misc{zhai_lit_2022,
	title = {{LiT}: {Zero}-{Shot} {Transfer} with {Locked}-image text {Tuning}},
	shorttitle = {{LiT}},
	url = {http://arxiv.org/abs/2111.07991},
	doi = {10.48550/arXiv.2111.07991},
	urldate = {2026-04-05},
	publisher = {arXiv},
	author = {Zhai, Xiaohua and Wang, Xiao and Mustafa, Basil and Steiner, Andreas and Keysers, Daniel and Kolesnikov, Alexander and Beyer, Lucas},
	month = jun,
	year = {2022},
	note = {arXiv:2111.07991 [cs]},
}

@misc{huang_act_2026,
	title = {Act {Like} a {Pathologist}: {Tissue}-{Aware} {Whole} {Slide} {Image} {Reasoning}},
	shorttitle = {Act {Like} a {Pathologist}},
	url = {http://arxiv.org/abs/2603.00667},
	doi = {10.48550/arXiv.2603.00667},
	urldate = {2026-04-05},
	publisher = {arXiv},
	author = {Huang, Wentao and Lyu, Weimin and Lou, Peiliang and Hu, Qingqiao and Hu, Xiaoling and Abousamra, Shahira and Han, Wenchao and Guo, Ruifeng and Zhou, Jiawei and Chen, Chao and Wang, Chen},
	month = mar,
	year = {2026},
	note = {arXiv:2603.00667 [cs]},
}

@article{ding_multimodal_2025,
	title = {A multimodal whole-slide foundation model for pathology},
	volume = {31},
	copyright = {2025 The Author(s)},
	issn = {1546-170X},
	url = {https://www.nature.com/articles/s41591-025-03982-3},
	doi = {10.1038/s41591-025-03982-3},
	language = {en},
	number = {11},
	urldate = {2026-04-05},
	journal = {Nature Medicine},
	publisher = {Nature Publishing Group},
	author = {Ding, Tong and Wagner, Sophia J. and Song, Andrew H. and Chen, Richard J. and Lu, Ming Y. and Zhang, Andrew and Vaidya, Anurag J. and Jaume, Guillaume and Shaban, Muhammad and Kim, Ahrong and Williamson, Drew F. K. and Robertson, Harry and Chen, Bowen and Almagro-Pérez, Cristina and Doucet, Paul and Sahai, Sharifa and Chen, Chengkuan and Chen, Christina S. and Komura, Daisuke and Kawabe, Akihiro and Ochi, Mieko and Sato, Shinya and Yokose, Tomoyuki and Miyagi, Yohei and Ishikawa, Shumpei and Gerber, Georg and Peng, Tingying and Le, Long Phi and Mahmood, Faisal},
	month = nov,
	year = {2025},
	pages = {3749--3761},
}

@inproceedings{wu_learning_2025,
	title = {Learning {Heterogeneous} {Tissues} with {Mixture} of {Experts} for {Gigapixel} {Whole} {Slide} {Images}},
	url = {https://openaccess.thecvf.com/content/CVPR2025/html/Wu_Learning_Heterogeneous_Tissues_with_Mixture_of_Experts_for_Gigapixel_Whole_CVPR_2025_paper.html},
	language = {en},
	urldate = {2025-11-23},
	author = {Wu, Junxian and Chen, Minheng and Ke, Xinyi and Xun, Tianwang and Jiang, Xiaoming and Zhou, Hongyu and Shao, Lizhi and Kong, Youyong},
	year = {2025},
	pages = {5144--5153},
}

@inproceedings{shao_multiple_2025,
	title = {Do {Multiple} {Instance} {Learning} {Models} {Transfer}?},
	url = {https://openreview.net/forum?id=hfLqdquVt3},
	language = {en},
	urldate = {2025-10-17},
	author = {Shao, Daniel and Chen, Richard J. and Song, Andrew H. and Runevic, Joel and Lu, Ming Y. and Ding, Tong and Mahmood, Faisal},
	month = jun,
	year = {2025},
}

@article{xu_whole-slide_2024,
	title = {A whole-slide foundation model for digital pathology from real-world data},
	volume = {630},
	issn = {0028-0836, 1476-4687},
	url = {https://www.nature.com/articles/s41586-024-07441-w},
	doi = {10.1038/s41586-024-07441-w},
	language = {en},
	number = {8015},
	urldate = {2025-09-20},
	journal = {Nature},
	author = {Xu, Hanwen and Usuyama, Naoto and Bagga, Jaspreet and Zhang, Sheng and Rao, Rajesh and Naumann, Tristan and Wong, Cliff and Gero, Zelalem and González, Javier and Gu, Yu and Xu, Yanbo and Wei, Mu and Wang, Wenhui and Ma, Shuming and Wei, Furu and Yang, Jianwei and Li, Chunyuan and Gao, Jianfeng and Rosemon, Jaylen and Bower, Tucker and Lee, Soohee and Weerasinghe, Roshanthi and Wright, Bill J. and Robicsek, Ari and Piening, Brian and Bifulco, Carlo and Wang, Sheng and Poon, Hoifung},
	month = jun,
	year = {2024},
	pages = {181--188},
}

@article{lu_visual-language_2024,
	title = {A visual-language foundation model for computational pathology},
	volume = {30},
	issn = {1078-8956, 1546-170X},
	url = {https://www.nature.com/articles/s41591-024-02856-4},
	doi = {10.1038/s41591-024-02856-4},
	language = {en},
	number = {3},
	urldate = {2025-09-20},
	journal = {Nature Medicine},
	author = {Lu, Ming Y. and Chen, Bowen and Williamson, Drew F. K. and Chen, Richard J. and Liang, Ivy and Ding, Tong and Jaume, Guillaume and Odintsov, Igor and Le, Long Phi and Gerber, Georg and Parwani, Anil V. and Zhang, Andrew and Mahmood, Faisal},
	month = mar,
	year = {2024},
	pages = {863--874},
}

@article{ehteshami_bejnordi_diagnostic_2017,
	title = {Diagnostic {Assessment} of {Deep} {Learning} {Algorithms} for {Detection} of {Lymph} {Node} {Metastases} in {Women} {With} {Breast} {Cancer}},
	volume = {318},
	issn = {0098-7484},
	url = {https://doi.org/10.1001/jama.2017.14585},
	doi = {10.1001/jama.2017.14585},
	number = {22},
	urldate = {2022-05-07},
	journal = {JAMA},
	author = {Ehteshami Bejnordi, Babak and Veta, Mitko and Johannes van Diest, Paul and van Ginneken, Bram and Karssemeijer, Nico and Litjens, Geert and van der Laak, Jeroen A. W. M. and {and the CAMELYON16 Consortium}},
	month = dec,
	year = {2017},
	pages = {2199--2210},
}

@article{lu_data-efficient_2021,
	title = {Data-efficient and weakly supervised computational pathology on whole-slide images},
	volume = {5},
	copyright = {2021 The Author(s), under exclusive licence to Springer Nature Limited},
	issn = {2157-846X},
	url = {https://www.nature.com/articles/s41551-020-00682-w},
	doi = {10.1038/s41551-020-00682-w},
	language = {en},
	number = {6},
	urldate = {2023-06-27},
	journal = {Nature Biomedical Engineering},
	author = {Lu, Ming Y. and Williamson, Drew F. K. and Chen, Tiffany Y. and Chen, Richard J. and Barbieri, Matteo and Mahmood, Faisal},
	month = jun,
	year = {2021},
	pages = {555--570},
}

@misc{shao_transmil_2021,
	title = {{TransMIL}: {Transformer} based {Correlated} {Multiple} {Instance} {Learning} for {Whole} {Slide} {Image} {Classification}},
	shorttitle = {{TransMIL}},
	url = {http://arxiv.org/abs/2106.00908},
	urldate = {2023-11-13},
	publisher = {arXiv},
	author = {Shao, Zhuchen and Bian, Hao and Chen, Yang and Wang, Yifeng and Zhang, Jian and Ji, Xiangyang and Zhang, Yongbing},
	month = oct,
	year = {2021},
}

@misc{song_artificial_2023,
	title = {Artificial {Intelligence} for {Digital} and {Computational} {Pathology}},
	url = {http://arxiv.org/abs/2401.06148},
	urldate = {2024-11-13},
	publisher = {arXiv},
	author = {Song, Andrew H. and Jaume, Guillaume and Williamson, Drew F. K. and Lu, Ming Y. and Vaidya, Anurag and Miller, Tiffany R. and Mahmood, Faisal},
	month = dec,
	year = {2023},
}

@misc{vorontsov_virchow_2024,
	title = {Virchow: {A} {Million}-{Slide} {Digital} {Pathology} {Foundation} {Model}},
	shorttitle = {Virchow},
	url = {http://arxiv.org/abs/2309.07778},
	urldate = {2024-11-14},
	publisher = {arXiv},
	author = {Vorontsov, Eugene and Bozkurt, Alican and Casson, Adam and Shaikovski, George and Zelechowski, Michal and Liu, Siqi and Severson, Kristen and Zimmermann, Eric and Hall, James and Tenenholtz, Neil and Fusi, Nicolo and Mathieu, Philippe and Eck, Alexander van and Lee, Donghun and Viret, Julian and Robert, Eric and Wang, Yi Kan and Kunz, Jeremy D. and Lee, Matthew C. H. and Bernhard, Jan and Godrich, Ran A. and Oakley, Gerard and Millar, Ewan and Hanna, Matthew and Retamero, Juan and Moye, William A. and Yousfi, Razik and Kanan, Christopher and Klimstra, David and Rothrock, Brandon and Fuchs, Thomas J.},
	month = jan,
	year = {2024},
}

@misc{hu_lora_2021,
	title = {{LoRA}: {Low}-{Rank} {Adaptation} of {Large} {Language} {Models}},
	shorttitle = {{LoRA}},
	url = {http://arxiv.org/abs/2106.09685},
	urldate = {2024-11-15},
	publisher = {arXiv},
	author = {Hu, Edward J. and Shen, Yelong and Wallis, Phillip and Allen-Zhu, Zeyuan and Li, Yuanzhi and Wang, Shean and Wang, Lu and Chen, Weizhu},
	month = oct,
	year = {2021},
}

@misc{li_dual-stream_2021,
	title = {Dual-stream {Multiple} {Instance} {Learning} {Network} for {Whole} {Slide} {Image} {Classification} with {Self}-supervised {Contrastive} {Learning}},
	url = {http://arxiv.org/abs/2011.08939},
	doi = {10.48550/arXiv.2011.08939},
	urldate = {2025-01-13},
	publisher = {arXiv},
	author = {Li, Bin and Li, Yin and Eliceiri, Kevin W.},
	month = apr,
	year = {2021},
}

@misc{ilse_attention-based_2018,
	title = {Attention-based {Deep} {Multiple} {Instance} {Learning}},
	url = {http://arxiv.org/abs/1802.04712},
	doi = {10.48550/arXiv.1802.04712},
	urldate = {2025-01-13},
	publisher = {arXiv},
	author = {Ilse, Maximilian and Tomczak, Jakub M. and Welling, Max},
	month = jun,
	year = {2018},
}

@article{xiang_visionlanguage_2025,
	title = {A vision–language foundation model for precision oncology},
	volume = {638},
	copyright = {2025 The Author(s), under exclusive licence to Springer Nature Limited},
	issn = {1476-4687},
	url = {https://www.nature.com/articles/s41586-024-08378-w},
	doi = {10.1038/s41586-024-08378-w},
	language = {en},
	number = {8051},
	urldate = {2025-03-01},
	journal = {Nature},
	author = {Xiang, Jinxi and Wang, Xiyue and Zhang, Xiaoming and Xi, Yinghua and Eweje, Feyisope and Chen, Yijiang and Li, Yuchen and Bergstrom, Colin and Gopaulchan, Matthew and Kim, Ted and Yu, Kun-Hsing and Willens, Sierra and Olguin, Francesca Maria and Nirschl, Jeffrey J. and Neal, Joel and Diehn, Maximilian and Yang, Sen and Li, Ruijiang},
	month = feb,
	year = {2025},
	pages = {769--778},
}

@misc{radford_learning_2021,
	title = {Learning {Transferable} {Visual} {Models} {From} {Natural} {Language} {Supervision}},
	url = {http://arxiv.org/abs/2103.00020},
	doi = {10.48550/arXiv.2103.00020},
	urldate = {2025-05-26},
	publisher = {arXiv},
	author = {Radford, Alec and Kim, Jong Wook and Hallacy, Chris and Ramesh, Aditya and Goh, Gabriel and Agarwal, Sandhini and Sastry, Girish and Askell, Amanda and Mishkin, Pamela and Clark, Jack and Krueger, Gretchen and Sutskever, Ilya},
	month = feb,
	year = {2021},
}

@article{chen_towards_2024,
	title = {Towards a general-purpose foundation model for computational pathology},
	volume = {30},
	copyright = {2024 The Author(s), under exclusive licence to Springer Nature America, Inc.},
	issn = {1546-170X},
	url = {https://www.nature.com/articles/s41591-024-02857-3},
	doi = {10.1038/s41591-024-02857-3},
	language = {en},
	number = {3},
	urldate = {2024-11-13},
	journal = {Nature Medicine},
	publisher = {Nature Publishing Group},
	author = {Chen, Richard J. and Ding, Tong and Lu, Ming Y. and Williamson, Drew F. K. and Jaume, Guillaume and Song, Andrew H. and Chen, Bowen and Zhang, Andrew and Shao, Daniel and Shaban, Muhammad and Williams, Mane and Oldenburg, Lukas and Weishaupt, Luca L. and Wang, Judy J. and Vaidya, Anurag and Le, Long Phi and Gerber, Georg and Sahai, Sharifa and Williams, Walt and Mahmood, Faisal},
	month = mar,
	year = {2024},
	pages = {850--862},
}

@article{lipkova_deep_2022,
	title = {Deep learning-enabled assessment of cardiac allograft rejection from endomyocardial biopsies},
	volume = {28},
	copyright = {2022 The Author(s), under exclusive licence to Springer Nature America, Inc.},
	issn = {1546-170X},
	url = {https://www.nature.com/articles/s41591-022-01709-2},
	doi = {10.1038/s41591-022-01709-2},
	language = {en},
	number = {3},
	urldate = {2022-05-10},
	journal = {Nature Medicine},
	author = {Lipkova, Jana and Chen, Tiffany Y. and Lu, Ming Y. and Chen, Richard J. and Shady, Maha and Williams, Mane and Wang, Jingwen and Noor, Zahra and Mitchell, Richard N. and Turan, Mehmet and Coskun, Gulfize and Yilmaz, Funda and Demir, Derya and Nart, Deniz and Basak, Kayhan and Turhan, Nesrin and Ozkara, Selvinaz and Banz, Yara and Odening, Katja E. and Mahmood, Faisal},
	month = mar,
	year = {2022},
	pages = {575--582},
}

@article{weinstein_cancer_2013,
	title = {The {Cancer} {Genome} {Atlas} {Pan}-{Cancer} {Analysis} {Project}},
	volume = {45},
	issn = {1061-4036},
	url = {https://www.ncbi.nlm.nih.gov/pmc/articles/PMC3919969/},
	doi = {10.1038/ng.2764},
	number = {10},
	urldate = {2022-03-31},
	journal = {Nature genetics},
	author = {Weinstein, John N. and Collisson, Eric A. and Mills, Gordon B. and Shaw, Kenna M. and Ozenberger, Brad A. and Ellrott, Kyle and Shmulevich, Ilya and Sander, Chris and Stuart, Joshua M.},
	month = oct,
	year = {2013},
	pages = {1113--1120},
}

@misc{qwen3technicalreport,
      title={Qwen3 Technical Report}, 
      author={Qwen Team},
      year={2025},
      eprint={2505.09388},
      archivePrefix={arXiv},
      primaryClass={cs.CL},
      url={https://arxiv.org/abs/2505.09388}, 
}

@inproceedings{li_wikg_2024,
  title={Dynamic graph representation with knowledge-aware attention for histopathology whole slide image analysis},
  author={Li, Jiawen and Chen, Yuxuan and Chu, Hongbo and Sun, Qiehe and Guan, Tian and Han, Anjia and He, Yonghong},
  booktitle={Proceedings of the IEEE/CVF conference on computer vision and pattern recognition},
  pages={11323--11332},
  year={2024}
}

@article{belagali2025ticon,
  title={TICON: A Slide-Level Tile Contextualizer for Histopathology Representation Learning},
  author={Belagali, Varun and Kapse, Saarthak and Marza, Pierre and Das, Srijan and Li, Zilinghan and Boutaj, Sofi{\`e}ne and Pati, Pushpak and Yellapragada, Srikar and Nandi, Tarak Nath and Madduri, Ravi K and others},
  journal={arXiv preprint arXiv:2512.21331},
  year={2025}
}

@article{yu2022coca,
title={CoCa: Contrastive Captioners are Image-Text Foundation Models},
author={Jiahui Yu and Zirui Wang and Vijay Vasudevan and Legg Yeung and Mojtaba Seyedhosseini and Yonghui Wu},
journal={Transactions on Machine Learning Research},
issn={2835-8856},
year={2022},
url={https://openreview.net/forum?id=Ee277P3AYC},
}

@inproceedings{li2023blip,
  title={Blip-2: Bootstrapping language-image pre-training with frozen image encoders and large language models},
  author={Li, Junnan and Li, Dongxu and Savarese, Silvio and Hoi, Steven},
  booktitle={International conference on machine learning},
  pages={19730--19742},
  year={2023},
  organization={PMLR}
}

@inproceedings{gorti2022x,
  title={X-pool: Cross-modal language-video attention for text-video retrieval},
  author={Gorti, Satya Krishna and Vouitsis, No{\"e}l and Ma, Junwei and Golestan, Keyvan and Volkovs, Maksims and Garg, Animesh and Yu, Guangwei},
  booktitle={Proceedings of the IEEE/CVF conference on computer vision and pattern recognition},
  pages={5006--5015},
  year={2022}
}

@inproceedings{shao_mixture_2025_v2,
	title = {Mixture of {Mini} {Experts}: {Overcoming} the {Linear} {Layer} {Bottleneck} in {Multiple} {Instance} {Learning}},
	shorttitle = {Mixture of {Mini} {Experts}},
	url = {https://openreview.net/forum?id=S5Io33pc78},
	language = {en},
	urldate = {2026-04-05},
	author = {Shao, Daniel and Runevic, Joel and Chen, Richard J. and Williamson, Drew FK and Kim, Ahrong and Song, Andrew H. and Mahmood, Faisal},
	month = oct,
	year = {2025},
}

@article{chen2024towards,
  title={Towards a general-purpose foundation model for computational pathology},
  author={Chen, Richard J and Ding, Tong and Lu, Ming Y and Williamson, Drew FK and Jaume, Guillaume and Song, Andrew H and Chen, Bowen and Zhang, Andrew and Shao, Daniel and Shaban, Muhammad and others},
  journal={Nature medicine},
  volume={30},
  number={3},
  pages={850--862},
  year={2024},
  publisher={Nature Publishing Group US New York}
}
\bibliographystyle{unsrtnat}

%%%%%%%%%%%%%%%%%%%%%%%%%%%%%%%%%%%%%%%%%%%%%%%%%%%%%%%%%%%%

\appendix
\newpage

\section{Appendix}
%%%%%%%%%%%%%%%%%%%%%%%%%%%%%%%%%%%%%%%%%%%%%%%%%%%%%%%%%%%%
\setcounter{table}{0}% reset table counter
\renewcommand{\thetable}{A\arabic{table}}% tables will be A1, A2, ...

\setcounter{figure}{0}% reset figure counter
\renewcommand{\thefigure}{A\arabic{figure}}% figures will be A1, A2, ...

% later in the document, insert the figure

\subsection{Supervised Implementation}
\label{sec:supervised_implementation}
We follow the standard training recipe for MIL models, as described in \cite{chen_towards_2024, shao_multiple_2025}. The experimental setup utilized the AdamW optimizer with a base learning rate of $1 \times 10^{-4}$ and a cosine annealing schedule. To optimize computational efficiency, we employed PyTorch’s native mixed-precision training. For benchmarks providing a validation split, models were trained for a maximum of 20 epochs, incorporating an early stopping mechanism with a 5-epoch patience window and a 10-epoch minimum threshold. The optimization objective was cross-entropy loss, paired with a random class-weighted sampling strategy and a batch size of 1. To prevent overfitting, we applied a weight decay of $1 \times 10^{-5}$ alongside a dropout rate of 0.25 within all feedforward layers and 0.1 for features extracted from the frozen encoder. 
% Training routines were conducted on three NVIDIA RTX 3090 GPUs.
Training and evaluation is performed on three NVIDIA RTX 3090 GPUs. Training for three epochs on the full training set was completed in 31 hours.

% =============================================================================
% APPENDIX: Dataset and Baseline Model Descriptions (v2 - verified)
% =============================================================================
% All factual claims verified against primary sources:
%  - PLIP:    Huang et al., Nature Medicine 2023 (10.1038/s41591-023-02504-3)
%  - Quilt:   Ikezogwo et al., NeurIPS 2023 (arXiv:2306.11207)
%  - MUSK:    Xiang et al., Nature 2025 (10.1038/s41586-024-08378-w)
%  - MI-Zero: Lu et al., CVPR 2023
%  - CONCH:   Lu et al., Nature Medicine 2024 (10.1038/s41591-024-02856-4)
%  - TITAN:   Ding et al., Nature Medicine 2025 (10.1038/s41591-025-03982-3)
%  - PRISM:   Shaikovski et al., arXiv:2405.10254 (and Paige HF model card)
%  - Datasets: original publications and grand-challenge / TCIA pages
% =============================================================================

\subsection{Dataset Descriptions}
\label{sec:appendix_dataset_full}

\paragraph{Internal Training Dataset.}
Our training corpus consists of \nslides H\&E and IHC stained WSIs paired with their corresponding diagnostic surgical pathology reports, sourced retrospectively from the digital pathology archives of two Massachusetts General Hospital and the Brigham and Women's Hospital under institutional review board approval. The dataset consists of the same slides utilized in TITAN\cite{ding_multimodal_2025}.
The corpus spans 20 organs and a broad mix of neoplastic, tissue damage, inflammatory, and normal cases. No TCGA, CAMELYON, HistAI/SPIDER, or CATCH slides/reports are included in training.

No additional manual annotation was performed on the slides or reports for our training: the supervisory signal is the routine clinical pathology report associated with each case at the time of original sign-out by the reporting pathologist. All WSIs and reports were de-identified prior to any computational analysis. Reports were further processed with a locally-hosted Qwen3-14B instance to remove residual identifying information and to filter content unrelated to histology, using the prompts provided in \textbf{Table \ref{tab:report_cleaning_prompt}}.

\begin{table}[htpb]
    \centering
    \caption{Prompt used to normalize pathology reports in the internal training dataset. The placeholder \texttt{\{report\}} denotes the raw pathology report text.}
    \label{tab:report_cleaning_prompt}
    \begin{tabular}{p{0.12\linewidth}p{0.78\linewidth}}
        \toprule
        \textbf{Role} & \textbf{Prompt content} \\
        \midrule
        System &
        You are a pathology text normalizer. Extract the diagnostic information from the text. Remove extraneous details not visible from histology, such as laterality, staining, immunohistochemistry, marker names, and positivity/negativity of genetic testing. Return the cleaned diagnosis text. \\
        \midrule
        User &
        Clean this pathology report. \newline\newline
        Report: \texttt{\{report\}} \newline\newline
        Example: \newline\newline
        Input: Metastatic poorly-differentiated adenocarcinoma with giant cell features identified in the liver, positive for CK7, GATA3, and focal p40, with PD-L1 expression in tumor cells. \newline\newline
        Output: Metastatic poorly-differentiated adenocarcinoma with giant cell features \\
        \bottomrule
    \end{tabular}
\end{table}

\paragraph{TCGA-OT.}
The TCGA-OT dataset \cite{ding_multimodal_2025} comprises 11,186 diagnostic H\&E formalin-fixed paraffin-embedded (FFPE) WSIs of primary tumors drawn from The Cancer Genome Atlas (TCGA) \cite{weinstein_cancer_2013}, a public consortium that aggregated tissue specimens contributed by tissue source sites at U.S. and Canadian cancer centers under the auspices of the National Cancer Institute and National Human Genome Research Institute. Each slide was assigned to one of 46 OncoTree disease subtypes by mapping the original TCGA clinical diagnosis (assigned by the contributing institution's pathologists at the time of sign-out) to the standardized OncoTree ontology, with the constraint that each retained class contain at least 50 samples using the splits provided by TITAN~\cite{ding_multimodal_2025}. IHC preparations, frozen sections, slides missing magnification metadata, samples with insufficient tumor, and slides from recurrent or metastatic growths were excluded. During quality control, we found that the OncoTree codes for melanoma (MEL) and cutaneous melanoma (SKCM) were not applied consistently across TCGA: both comprise predominantly cutaneous melanomas, and the assignment appears to reflect whether a subtype was recorded at sign-out rather than a histological distinction. We therefore merged the two into a single melanoma class, yielding 45 classes. The dataset is partitioned into 8,226 training, 1,612 validation, and 1,348 test slides.

Rather than a 45-way tissue-agnostic classification, we formulate this as an organ-specific multiple-choice task, mirroring clinical workflows where tissue origin is known \textit{a priori}. For each WSI, valid diagnostic options are restricted to its corresponding organ. To account for 11 of the 28 organ codes containing only a single diagnostic label, we group bladder (n=457), prostate (n=449), cervix (n=229), and the remaining cases (n=409) to a single category. For reporting concision, we report the 7 organs with the highest volume of slides, and group the remaining into an "Other" category, reported as macro-averaged balanced accuracy.

\paragraph{SlideChat-HistAI.}

We use a HistAI-derived multiple-choice subset of the SlideChat evaluation suite, comprising histological subtyping over 1,396 WSIs \cite{chen_slidechat_2025}. The underlying slides are drawn from the HistAI open-source whole-slide image dataset \cite{nechaev_histai_2025}, an aggregation of de-identified diagnostic WSIs scanned predominantly on Leica Aperio GT450 and AT2 systems and curated by HistAI, accompanied by structured clinical metadata (diagnosis, demographics, standardized diagnostic codes). While originally designed for LLM-based question answering, the same multiple choice options are provided for all slides within an organ. Consequently, we treat this as a three-way zero-shot classification task and report balanced accuracy within each organ. 

\paragraph{SPIDER.}
The Supervised Pathology Image-DEscription Repository (SPIDER) \cite{nechaev_spider_nodate} is a patch-level multi-organ benchmark released by HistAI, with WSIs sourced from the broader HistAI WSI dataset \cite{nechaev_histai_2025}. We evaluate on the four publicly released organ subsets: skin (159,854 patches, 24 classes), breast (92,892 patches, 18 classes), colorectal (77,182 patches, 13 classes), and thorax (78,307 patches, 14 classes). Annotations were generated by board-certified pathologists using HistAI's web-based CELLDX annotation platform: for each candidate slide, pathologists delineated regions of the target morphology with polygon and brush tools, and the annotated regions were tessellated into non-overlapping $224 \times 224$ pixel patches at $20\times$ magnification. Each released patch is paired with surrounding context patches and a label verified by an expert pathologist. We report both fine-grained classification (SPIDER-F) using the official class labels and a unified 7-class mapping (SPIDER-C: Tumor, Stroma, Inflammation, Necrosis, Epithelium, Vascular, Fat), with additional categories, Nerve and Skin adnexa (Skin) and Mucin (Colorectal), defined in consultation with an in-house pathologist.

\paragraph{CAMELYON-16 and CAMELYON+.}
CAMELYON-16 \cite{ehteshami_bejnordi_diagnostic_2017} consists of 400 H\&E-stained WSIs of sentinel lymph node sections from breast cancer patients, collected at Radboud University Medical Center (Nijmegen, the Netherlands) and the University Medical Center Utrecht (Utrecht, the Netherlands). The original ground truth was produced by pathologists at the contributing institutions, who delineated metastatic regions on each tumor-positive slide as polygon contours stored in XML and binary mask formats; non-tumor slides were verified as negative. Because the original CAMELYON-16 annotations are known to contain errors and incomplete pixel-level coverage, we use the CAMELYON+ re-annotation by Ling \textit{et al.} \cite{ling_comprehensive_2025}, which jointly reprocesses the CAMELYON-16 and CAMELYON-17 collections: a panel of expert pathologists removed low-quality slides, corrected erroneous slide-level labels, and provided dense pixel-level tumor annotations including for slides previously lacking them. From the CAMELYON+ pixel-level masks we derive patch-level labels of metastasis vs. non-cancerous lymph node tissue, yielding 206,800 patches across 48 slides (71,564 with metastasis).

\paragraph{CATCH.}
The Pan-Tumor Canine Cutaneous Cancer Histology (CATCH) dataset \cite{wilm_pan-tumor_2022} consists of 350 H\&E-stained WSIs of cutaneous tissue biopsies from 282 canine patients, retrospectively selected from the biopsy archive of the Institute for Veterinary Pathology, Freie Universit\"at Berlin. Slides were stained with routine hematoxylin and eosin and digitized on Leica linear scanning systems at $0.25\,\mu m/\text{pixel}$. The dataset includes 12{,}424 polygon annotations across 13 classes: six non-neoplastic tissue classes (epidermis, dermis, subcutis, bone, cartilage, and a joint inflammation/necrosis class) and seven cutaneous tumor subtypes (melanoma, mast cell tumor, squamous cell carcinoma, peripheral nerve sheath tumor, plasmacytoma, trichoblastoma, and histiocytoma). Annotations were drawn by veterinary pathologists using the open-source SlideRunner tool, and label quality was validated through inter-rater experiments with three independent pathologists, demonstrating high concordance particularly for tumor regions. We tessellate the annotated WSIs into patches and assign a patch the dominant class label only when at least 99\% of the patch area falls within a single annotated class; patches with conflicting labels are excluded. Under-represented tissues (bone, cartilage) are excluded following previous pipelines \cite{belagali2025ticon}, yielding 11 classes. Final patch counts are reported in Table~\ref{tab:catch_counts}.

% =============================================================================

\begin{table}[htpb]
    \centering
    \caption{Patch Counts by Label for CATCH dataset. A patch label was given a label if >99\% of the area within that patch consisted of that label. Patches without >99\% area containing a single label were not included for analysis (conflicting labels).}
    \label{tab:catch_counts}
    \begin{tabular}{lr}
        \toprule
        \textbf{Label} & \textbf{Count} \\
        \midrule
        Dermis & 263,813 \\
        Epidermis & 23,579 \\
        Histiocytoma & 50,770 \\
        Inflamm/Necrosis & 32,723 \\
        Mast Cell Tumor & 166,951 \\
        Melanoma & 118,835 \\
        PNST & 193,608 \\
        Plasmacytoma & 84,548 \\
        SCC & 63,839 \\
        Subcutis & 90,682 \\
        Trichoblastoma & 161,933 \\
        \midrule
        Conflicting labels & 1,978 \\
        \bottomrule
    \end{tabular}
\end{table}

\subsection{Baseline Model Descriptions}
\label{sec:appendix_models}

We evaluate \ours against six published vision-language pathology baselines spanning patch-level and slide-level granularities. We summarize the data scale and pretraining approach of each model in Table~\ref{tab:baseline_models}, and provide individual descriptions below.

\paragraph{PLIP \cite{huang_visuallanguage_2023}.}
Pathology Language-Image Pretraining (PLIP) is a CLIP-style vision-language model produced by fine-tuning OpenAI's CLIP (ViT-B/32 image encoder, transformer text encoder) on OpenPath, a corpus of 208,414 image-text pairs curated from pathology-related Twitter hashtags (collected between March 2006 and November 2022) and supplemented with pathology subsets of LAION-5B. Captions originate from the natural-language posts and replies of pathologists on social media, with quality filtering applied during curation. Training uses the standard symmetric image-text contrastive (InfoNCE) objective.

\paragraph{QuiltNet \cite{ikezogwo_quilt-1m_nodate}.}
QuiltNet is a CLIP fine-tuned on Quilt-1M, a corpus of approximately one million pathology image-text pairs. The Quilt subset (768K--802K pairs depending on release) is extracted from 1{,}087 hours of educational pathology videos across 4{,}504 YouTube videos using a pipeline that combines automatic speech recognition, GPT-3.5-based ASR error correction, and UMLS-grounded text post-processing to align narrated commentary with extracted video frames. The remainder of Quilt-1M is drawn from Twitter, research papers, and the open web (including OpenPath and ARCH). Training follows the standard CLIP contrastive objective. We use the QuiltNet-b16 variant for this study.

\paragraph{MUSK \cite{xiang_visionlanguage_2025}.}
MUSK (Multimodal transformer with Unified maSK modeling), developed by Stanford and Harvard, is a BEiT-3-based vision-language model with shared self-attention blocks and modality-specific feed-forward experts. The first pretraining stage performs masked modeling independently on 50 million pathology image patches (extracted from approximately 33{,}000 WSIs spanning 11{,}577 patients across 33 tumor types) and on 1 billion pathology-related text tokens, without requiring image-text pairing. The second stage performs CLIP-style contrastive alignment on the 1M paired Quilt-1M dataset \cite{ikezogwo_quilt-1m_nodate}, augmented with an auxiliary masked language modeling loss.

\paragraph{MI-Zero \cite{lu_visual_2023}.}
MI-Zero is a multiple instance learning framework for zero-shot WSI classification built on a CLIP-style patch encoder. The released encoders use CTransPath as the vision backbone and either BioClinicalBERT or PubMedBERT as the text encoder, fine-tuned via InfoNCE on a corpus of 33{,}480 patch-level image-caption pairs assembled from the ARCH dataset (7{,}562 pairs from educational material and PubMed figure legends) and additional internal data. At inference, slide-level predictions are obtained by computing patch-level cosine similarities to class prompts and aggregating via top-$K$ pooling. We use the BioClinicalBERT variant for this study. 

\paragraph{CONCH and CONCHv1.5 \cite{lu_visual-language_2024,ding_multimodal_2025}.}
CONCH (CONtrastive learning from Captions for Histopathology) is a CoCa-style vision-language patch encoder pretrained on 1.17 million human histopathology image-caption pairs curated from the PubMed Central Open Access (PMC OA) dataset and educational sources, with three-stage automated cleaning (histopathology image detection, caption splitting, and image-caption matching). The training objective combines contrastive image-text alignment with an auxiliary captioning loss. CONCHv1.5 upgrades the original CONCH vision-language patch encoder by integrating a Vision Transformer Large (ViT-L/16) backbone restored from the UNI checkpoint \cite{chen2024towards} and fine-tuning it under the same CoCa objective on a slightly larger and more diverse 1.26 million image-caption corpus that spans FFPE, frozen, and IHC tissue types. 

\paragraph{TITAN \cite{ding_multimodal_2025}.}
TITAN is a slide-level vision-language foundation model from \nslides medical reports from Mass General Brigham hospital, supplemented with publicly available GTEx slides for the vision-only pretraining stage. Pretraining proceeds in three stages: (1) vision-only iBOT-style self-supervised pretraining on $4 \times 4\,\text{mm}^2$ ROI crops, (2) CoCa-style contrastive alignment of slide embeddings with $\sim$423K synthetic ROI-level captions, and (3) CoCa-style alignment of full WSI embeddings with paired clinical reports. TITAN consumes patch features extracted by CONCHv1.5 at $20\times$ magnification.

\paragraph{PRISM \cite{shaikovski_prism_nodate}.}
PRISM (Pathology Report and Image Summarization Model) is a slide-level vision-language model pretrained on an internal corpus of 587{,}000 WSIs and 195{,}000 clinical pathology reports from Memorial Sloan Kettering Cancer Center; reports were rewritten using GPT-4 prior to training. PRISM consists of a Perceiver-based slide encoder operating on Virchow V1 patch tile embeddings and a BioGPT-based language decoder, jointly trained with the CoCa objective (contrastive image-text alignment plus autoregressive captioning). The model contains $\sim$558M parameters.

\subsection{Ablation Implementation}
\label{sec:ablation_losses}
\textbf{Sigmoid Locked-Image Tuning (SigLIT)}

Sigmoid Locked-Image Tuning (SigLIT) \cite{zhai_sigmoid_2023} is optimized by treating each slide-report pairing as an independent binary classification task. We define an indicator label $y_i \in \{-1, 1\}$, where $y_i = 1$ if $R_i$ is the true matching report for the slide, and $y_i = -1$ if it is a negative sample. The total loss for a given slide is the sum of the negative log-likelihoods over the $R$ sampled reports, computed using the sigmoid function $\sigma(\cdot)$:
\begin{equation}
    \mathcal{L} = - \frac{1}{R} \sum_{i=1}^R \log \sigma \left( y_i \left( \frac{\mathbf{h}_W^\top \mathbf{r}_i}{\tau} + b \right) \right)
\end{equation}
where $\mathbf{h}_W$ is the slide-level visual representation, $\tau$ is a learnable temperature scalar, and $b$ is a learnable bias term that shifts the logits before the sigmoid activation.

\begin{figure}[htbp]
    \centering
    \includegraphics[width=1\linewidth]{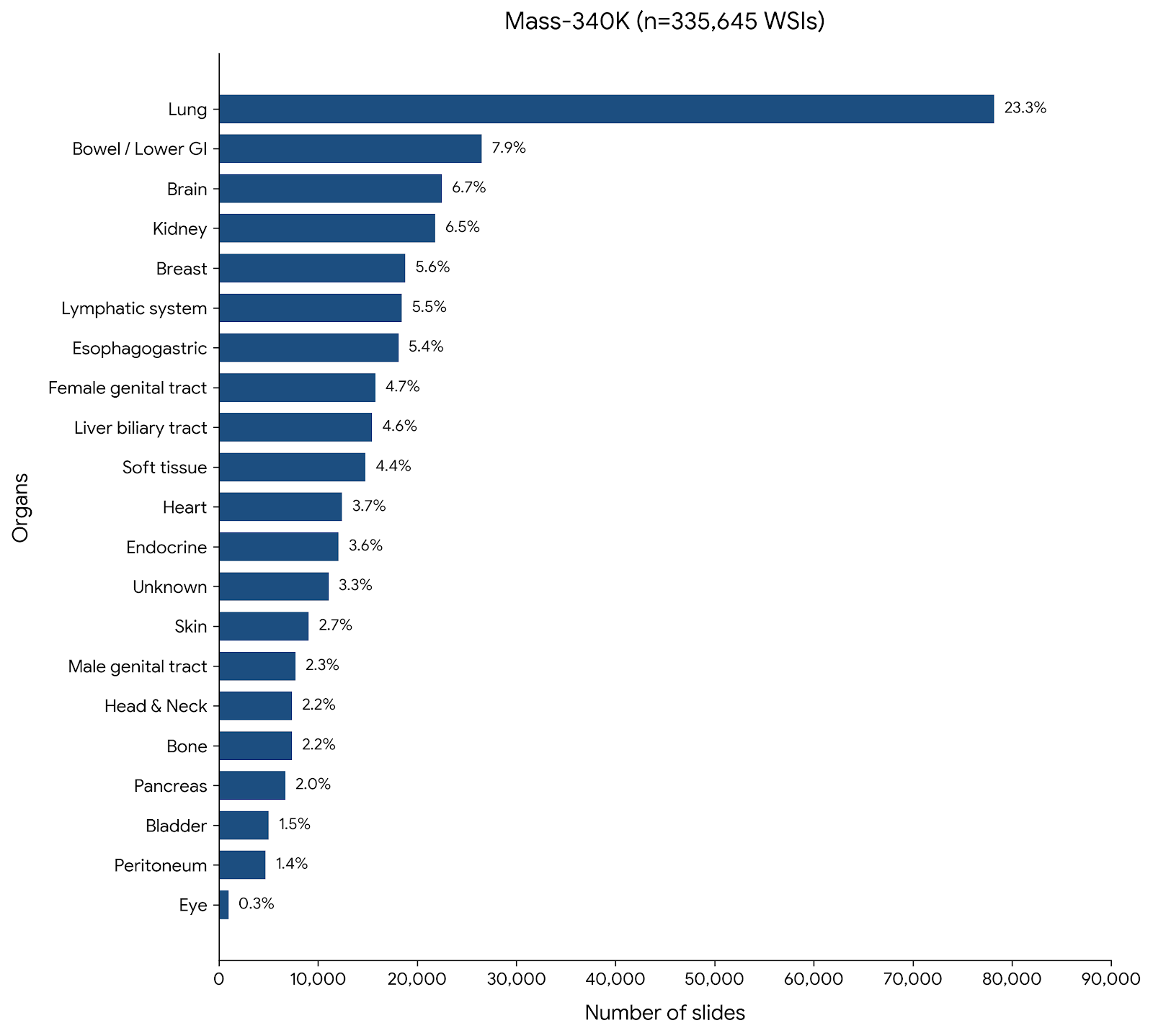}
    \caption{Training data distribution by organ.}
    \label{fig:training_distribution}
\end{figure}

\textbf{Training Sizes}: To obtain datasets at 150k and 50k samples, we perform stratified sampling without replacement to preserve the distribution of organs on the full pretraining set (\textbf{Figure \ref{fig:training_distribution}}). We evaluate the influence of inference-time ablations (mean pool, random pool) at the 150k and 50k scale using the models trained with the LIMIT approach.

\bigskip

\begin{table}[htbp]
\centering
\caption{Summary of baseline vision-language pathology models. ``Vision corpus'' reports the scale of unique image data (image-text pairs, patches, or WSIs) used for vision-language alignment. ``Reports/captions'' reports the scale of paired textual supervision when distinct from the vision corpus. ``Source institutions'' lists the primary clinical or public sources of training data; public corpora are denoted in italics. Citations refer to the model release papers.}
\label{tab:baseline_models}
\resizebox{\textwidth}{!}{
\begin{tabular}{lllllll}
\toprule
\textbf{Model} & \textbf{Architecture} & \textbf{Vision corpus} & \textbf{Reports/captions} & \textbf{Source institution(s)} & \textbf{Objective} & \textbf{Granularity} \\
\midrule
PLIP \cite{huang_visuallanguage_2023} & ViT-B/32 + Transformer & 208K pairs & (paired) & \textit{Twitter (\#path), LAION-5B} & CLIP InfoNCE & Patch \\
QuiltNet \cite{ikezogwo_quilt-1m_nodate} & ViT-B/32 + Transformer & 1M pairs & (paired) & \textit{YouTube, Twitter, web, ARCH} & CLIP InfoNCE & Patch \\
MI-Zero \cite{lu_visual_2023} & CTransPath + BioClinicalBERT & 33K pairs & (paired) & \textit{ARCH} + internal & InfoNCE + MIL pool & Patch / WSI \\
MUSK \cite{xiang_visionlanguage_2025} & BEiT-3 (multiway transformer) & 50M patches ($\sim$33K WSIs) + 1M pairs & 1B tokens (unpaired) & Stanford internal; \textit{Quilt-1M} & Masked modeling + CLIP & Patch \\
CONCHv1.5 \cite{lu_visual-language_2024,ding_multimodal_2025} & ViT (CoCa, UNI-init) & 1.17M pairs (vision encoder only released) & (paired) & \textit{PMC OA, EDU} & Contrastive + captioning & Patch \\
TITAN \cite{ding_multimodal_2025} & ViT slide encoder (iBOT + CoCa) & 335K WSIs + 423K ROI-captions & 183K reports & BWH, MGH; \textit{GTEx} & SSL + contrastive + captioning & WSI \\
PRISM \cite{shaikovski_prism_nodate} & Perceiver + BioGPT & 587K WSIs (Virchow tiles) & 195K reports & MSKCC & CoCa (contrastive + captioning) & WSI \\
\midrule
\ours (ours) & CONCHv1.5 (frozen) + LiT text encoder & 335K WSIs & 335K reports & BWH, MGH, \textit{GTEx} & Difficulty-weighted contrastive (LIMIT) & Patch / WSI \\
\bottomrule
\end{tabular}
}
\end{table}

\begin{figure}
    \centering
    \includegraphics[width=1\linewidth]{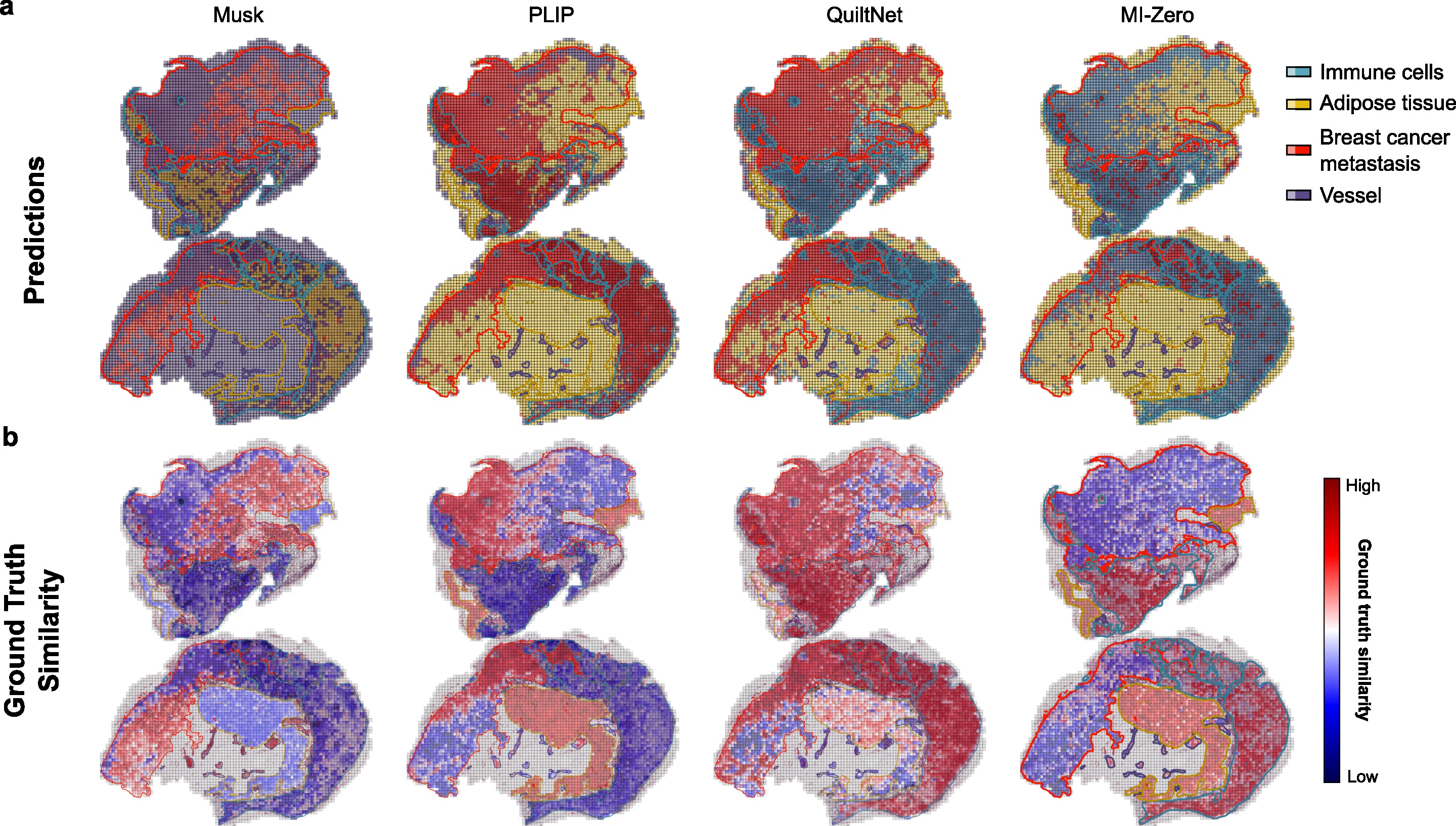}
    \caption{\textbf{Zero-shot breast cancer metastasis in lymph nodes for various vision-language models}. \textbf{a)} Patch-level predictions according to the class embedding with the highest cosine similarity to each patch embedding. Annotated regions according to an in-house pathologist are overlayed. \textbf{b)} Direct similarity scores between each class embedding and the patch embedding. More red indicates higher similarity. Scores are normalized within each model. Among the baseline patch models, QuiltNet exhibits the highest degree of overlap with the annotated regions.}
    \label{fig:app_heatmap}
\end{figure}

\begin{figure}
    \centering
    \includegraphics[width=1\linewidth]{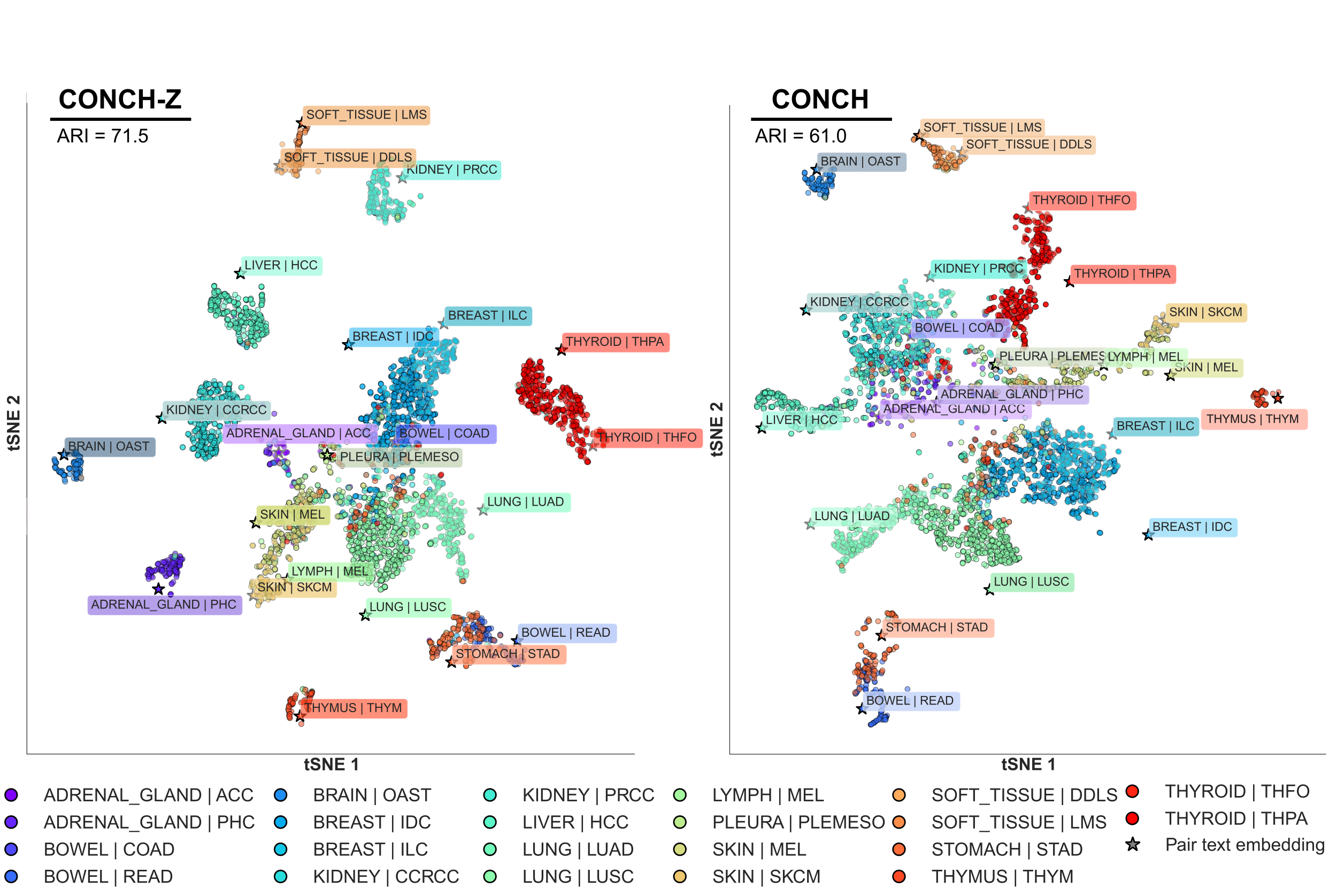}
    \caption{\textbf{Alignment of text embeddings with slide representations on TCGA-OT}. T-SNE on top 10 components of canonical correlation analysis of cancer subtypes and slide-level features. Slide-level features were obtained by weighted average with query text embedding "cancer". Textual embeddings of \ours exhibit higher overlap with image embeddings than CONCHv1.5 text embeddings. For instance, Adjusted rand index (ARI) indicates substantially better alignment for \ours (ARI=71.5) compared to CONCHv1.5 (ARI=61.0).}
    \label{fig:app_tsne}
\end{figure}

\begin{table}[htpb]
    \centering
    \caption{Terms and Prompt Templates for SPIDER-C. Synonyms are used as interchangeable terms, and templates define text prompt structures where \{term\} is replaced by each synonym.}
    \label{tab:spider_terms_templates}
    \begin{tabular}{ll}
        \toprule
        \textbf{Category} & \textbf{Content} \\
        \midrule
        Synonyms & Tumor; Cancer; Malignancy; Carcinoma; Neoplasm \\
        \midrule
        Templates & \{term\} \\
                  & an image of \{term\} \\
                  & \{term\} cells \\
        \bottomrule
    \end{tabular}
\end{table}

%\subsection{Broader Impacts}

%This work may benefit computational pathology by improving zero-shot tissue and slide-level classification, potentially reducing annotation burden and supporting exploratory analysis of whole-slide images. However, \ours is not validated for autonomous clinical diagnosis, and incorrect predictions could cause harm if used without pathologist oversight. Performance may vary across institutions, scanners, staining protocols, demographic groups, and rare disease subtypes. Any clinical use would require prospective validation, monitoring for distribution shift and bias, privacy-preserving data governance, and deployment only as an assistive tool under qualified medical supervision.

%\newpage

%\input{checklist.tex}

\end{document}